\documentclass{article}
\pdfoutput=1
\usepackage{microtype}
\usepackage{graphicx}
\usepackage{booktabs} 

\usepackage{hyperref}

\usepackage[accepted]{icml2023}

\usepackage{amsmath}
\usepackage{amssymb}
\usepackage{mathtools}
\usepackage{amsthm}
\usepackage{nicefrac}       %
\usepackage{amsmath}
\usepackage{tabularx}
\usepackage{inconsolata}

\usepackage{algorithmicx, algpseudocode}
\usepackage{subcaption}
\usepackage{multirow}
\usepackage[capitalize,noabbrev]{cleveref}

\theoremstyle{plain}

\theoremstyle{definition}

\theoremstyle{remark}

\definecolor{blue}{rgb}{0.3, 0.3, 0.8}
\definecolor{red}{rgb}{0.8, 0.3, 0.3}
\definecolor{green}{rgb}{0.3, 0.8, 0.3}

\newcommand{\red}[1] {\textcolor{red}{#1}}
\newcommand{\blue}[1] {\textcolor{blue}{#1}}

\newcommand{\rebuttal}[1] {{#1}}

\newcommand{\eg} {e.g.\,}

\usepackage[textsize=tiny]{todonotes}

\icmltitlerunning{Guiding Pretraining in Reinforcement Learning with Large Language Models}

\begin{document}

\twocolumn[
\icmltitle{Guiding Pretraining in Reinforcement Learning with Large Language Models}



\icmlsetsymbol{equal}{*}

\begin{icmlauthorlist}
\icmlauthor{Yuqing Du}{equal,yyy}
\icmlauthor{Olivia Watkins}{equal,yyy}
\icmlauthor{Zihan Wang}{comp}
\icmlauthor{Cédric Colas}{sch,inr}
\icmlauthor{Trevor Darrell}{yyy}
\icmlauthor{Pieter Abbeel}{yyy}
\icmlauthor{Abhishek Gupta}{comp}
\icmlauthor{Jacob Andreas}{sch}
\end{icmlauthorlist}

\icmlaffiliation{yyy}{Department of Electrical Engineering and Computer Science, University of California, Berkeley, USA}
\icmlaffiliation{comp}{University of Washington, Seattle}
\icmlaffiliation{sch}{Massachusetts Institute of Technology, Computer Science and Artificial Intelligence Laboratory}
\icmlaffiliation{inr}{Inria, Flowers Laboratory}

\icmlcorrespondingauthor{Yuqing Du}{yuqing\_du@berkeley.edu}
\icmlcorrespondingauthor{Olivia Watkins}{oliviawatkins@berkeley.edu}

\icmlkeywords{Machine Learning, ICML}

\vskip 0.3in
]



\printAffiliationsAndNotice{\icmlEqualContribution} 
\begin{abstract}
Reinforcement learning algorithms typically struggle in the absence of a dense, well-shaped reward function. 
Intrinsically motivated exploration methods address this limitation by rewarding agents for visiting novel states or transitions, but these methods offer limited benefits in large environments where most discovered novelty is irrelevant for downstream tasks. 
We describe a method that uses background knowledge from text corpora to shape exploration.
This method, called ELLM (\textbf{E}xploring with \textbf{LLM}s) rewards an agent for achieving goals suggested by a language model prompted with a description of the agent's current state.
By leveraging large-scale language model pretraining, ELLM guides agents toward human-meaningful and plausibly useful behaviors without requiring a human in the loop.
We evaluate ELLM in the \textit{Crafter} game environment and the \textit{Housekeep} robotic simulator,
showing that ELLM-trained agents have better coverage of common-sense behaviors during pretraining and usually match or improve performance on a range of downstream tasks. Code available at \url{https://github.com/yuqingd/ellm}. 
\end{abstract}

\section{Introduction}

Reinforcement learning algorithms work well when learners receive frequent rewards that incentivize progress toward target behaviors. But hand-defining such reward functions 
requires significant engineering efforts in all but the simplest cases \citep{amodei2016concrete,evo_creativity}. 
To master complex tasks in practice, RL agents may therefore need to learn some behaviors in the absence of externally-defined rewards. What should they learn?

\begin{figure}[t!]	
\centering
\includegraphics[width=\columnwidth]{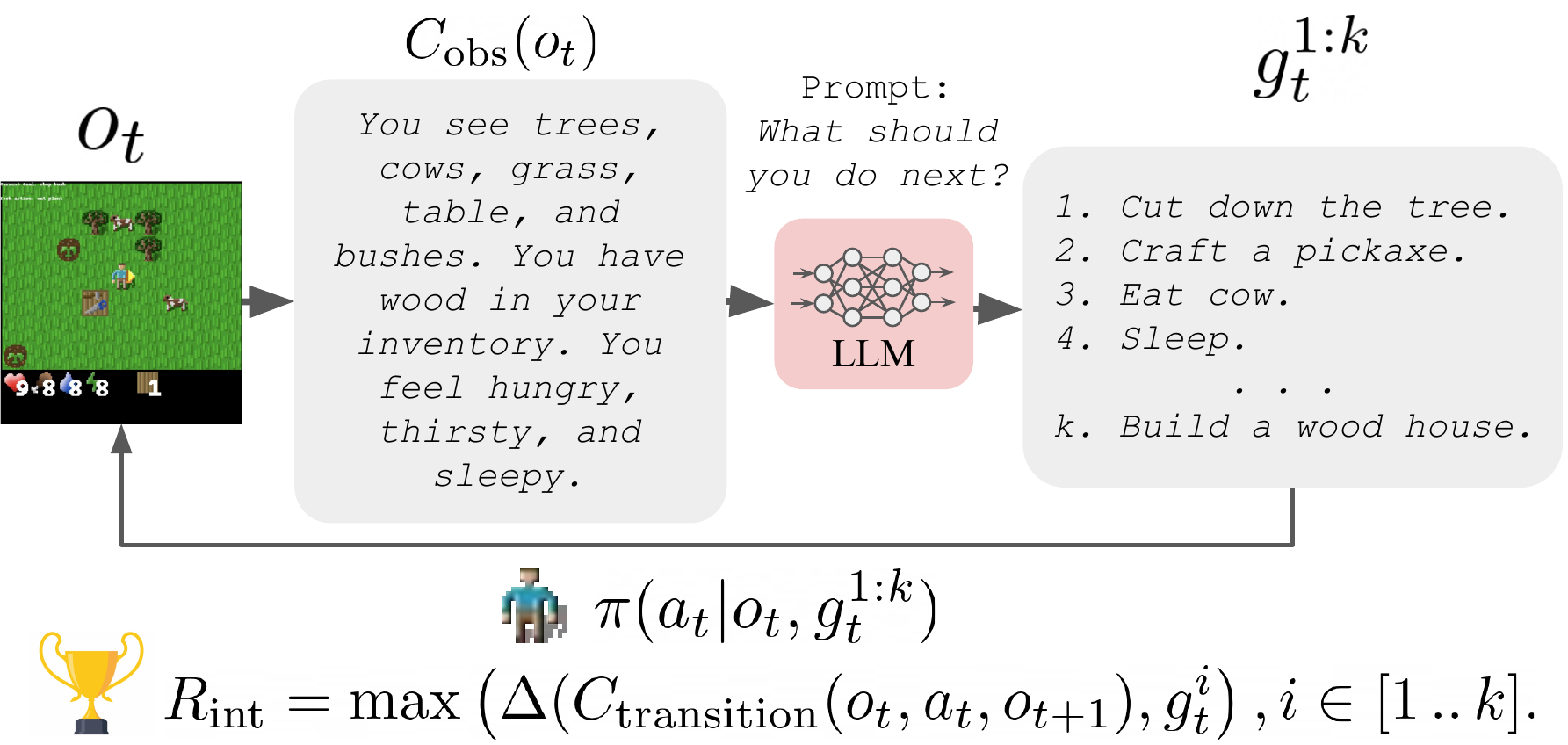}
\caption{\textbf{ELLM} uses a pretrained large language model (LLM) to suggest plausibly useful goals in a task-agnostic way. Building on LLM capabilities such as context-sensitivity and common-sense, ELLM trains RL agents to pursue goals that are likely meaningful without requiring direct human intervention. Prompt is illustrative; see full prompt and goal format in Appendix \ref{sec:crafter-prompt}. 
}
\label{fig:suggestion_pipeline}
\end{figure}

Intrinsically motivated RL methods answer this question by 
augmenting rewards
with auxiliary objectives based on novelty, surprise, uncertainty, or prediction errors \citep{bellemare2016unifying, pathak2017curiosity, burda2019exploration, zhang2021noveld, liu2021behavior, yarats2021reinforcement}. 
But not everything novel or unpredictable is useful: noisy TVs and the movements of leaves on a tree may provide an infinite amount of novelty, but do not lead to meaningful behaviors \citep{burda2019exploration}. More recent approaches compute novelty with higher-level representations like language \citep{tam2022semantic, mu2022improving}, but can continue driving the agent to explore behaviors that are unlikely to correspond to any human-meaningful goal---like enumerating unique configurations of furniture in a household.
It is not sufficient for extrinsic-reward-free RL agents to optimize for novelty alone: learned behaviors must also be useful. 

In this paper, we describe a method for using not just language-based representations but \textbf{pretrained language models} (LLMs) as a source of information about useful behavior.
LLMs are probabilistic models of text trained on large text corpora; their predictions encode rich information about human common-sense knowledge and cultural conventions.
Our method, \textbf{E}xploring with \textbf{LLM}s (ELLM),
queries LMs for possible goals given an agent's current context and rewards agents for accomplishing those suggestions. As a result, exploration is biased towards completion of goals that are diverse, context-sensitive, and human-meaningful. 
ELLM-trained agents exhibit better coverage of useful behaviors during pretraining, and outperform or match baselines when fine-tuned on downstream tasks. 


\section{Background and Related Work}


\paragraph{Intrinsically Motivated RL.} When reward functions are sparse, agents often need to carry out a long, specific sequence of actions to achieve target tasks. 
As action spaces or target behaviors grow more complex, the space of alternative action sequences agents can explore grows combinatorially. In such scenarios, 
undirected exploration that randomly perturbs actions or policy parameters has little chance of succeeding \citep{ten2022curiosity,ladosz2022exploration}.

Many distinct action sequences can lead to similar outcomes \citep{baranes2013active}---for example, most action sequences cause a humanoid agent to fall, while very few make it walk. Building on this observation, \textbf{intrinsically motivated} RL algorithms (IM-RL) choose to explore \emph{outcomes} rather than actions \citep{oudeyer2009intrinsic,ten2022curiosity,ladosz2022exploration}. \textbf{Knowledge-based} IMs (KB-IMs) focus on maximising the diversity of states
\citep[reviews in][]{aubret2019survey,linke2020adapting}. \textbf{Competence-based IMs} (CB-IMs) maximise the diversity of \emph{skills} mastered by the agent 
\citep[review in][]{colas2022autotelic}. Because most action sequences lead to a very restricted part of the outcome space (\eg all different ways of \textit{falling on the floor} likely correspond to a single outcome), these methods lead to a greater diversity of outcomes than undirected exploration \citep{lehman2008exploiting,colas2018gep}. 

However, maximizing diversity of outcomes may not always be enough. Complex environments can contain sources of infinite novelty. 
In such environments, seeking ever-more-novel states might drive learning towards behaviors that have little relevance to the true task reward. Humans do not explore outcome spaces uniformly, but instead rely on their physical and social common-sense to explore \textit{plausibly-useful} behaviors first. In video games, they know that keys should be used to open doors, ladders should be climbed, and snakes might be enemies. If this semantic information is removed, their exploration becomes severely impacted \citep{dubey2018investigating}. The approach we introduce in this paper, ELLM, may be interpreted as a CB-IM algorithm that seeks to explore the space of possible and plausibly-useful skills informed by human prior knowledge.

\paragraph{Linguistic Goals and Pretrained Language Models.}
One way of representing a diverse outcome space for exploration is through language. Training agents to achieve language goals brings several advantages: (1) goals are easy to express for non-expert users; (2) they can be more abstract than standard state-based goals \citep{colas2022autotelic}; and (3) agents can generalize better thanks to the partial compositionality and recursivity of language \citep{hermann2017grounded,hill2019environmental,colas2020language}. Such linguistic goals can be used as instructions for language-conditioned imitation learning or RL. In RL, agents typically receive language instructions corresponding to the relevant reward functions \citep{luketina2019survey} and are only rarely intrinsically motivated \citep[with the exception of][]{mu2022improving,colas2020language,tam2022semantic}, where language is also used as a more general compact state abstraction for task-agnostic exploration.

Representing goals in language unlocks the possibility of using text representations and generative models of text (large language models, or LLMs) trained on large corpora.
In imitation learning, text pretraining can help learners automatically recognize sub-goals and learn modular sub-policies from unlabelled demonstrations \citep{lynch2020language, sharma2021skill}, or chain pre-trained goal-oriented policies together to accomplish high-level tasks \citep{yao2020keep,huang2022language,saycan,huang2022inner}.
In RL, LM-encoded goal descriptions greatly improve the generalization of instruction-following agents across instructions \citep{chan2019actrce} and from synthetic to natural goals \citep{hill2020human}. LLMs have also been used as proxy reward functions when prompted with desired behaviors \citep{kwonreward}. Unlike these approaches, ELLM uses pretrained LLMs to constrain exploration towards plausibly-useful goals in a task-agnostic manner. It does not assume a pretrained low-level policy, demonstrations, or task-specific prompts. Most similar to our work, \citet{choi2022lmpriors} also prompt LLMs for priors. However, they use LM priors to classify safe and unsafe states to reward, which is a subset of common-sense exploratory behaviors ELLM should generate.  Also similar to our work, \citet{housekeep} query LLMs for zero-shot commonsense priors in the Housekeep environment, but they apply these to a planning task rather than as rewards for reinforcement learning.

\begin{figure*}[ht!]
\centering
\begin{subfigure}{.6\textwidth}
\centering
\includegraphics[width=.9\textwidth]{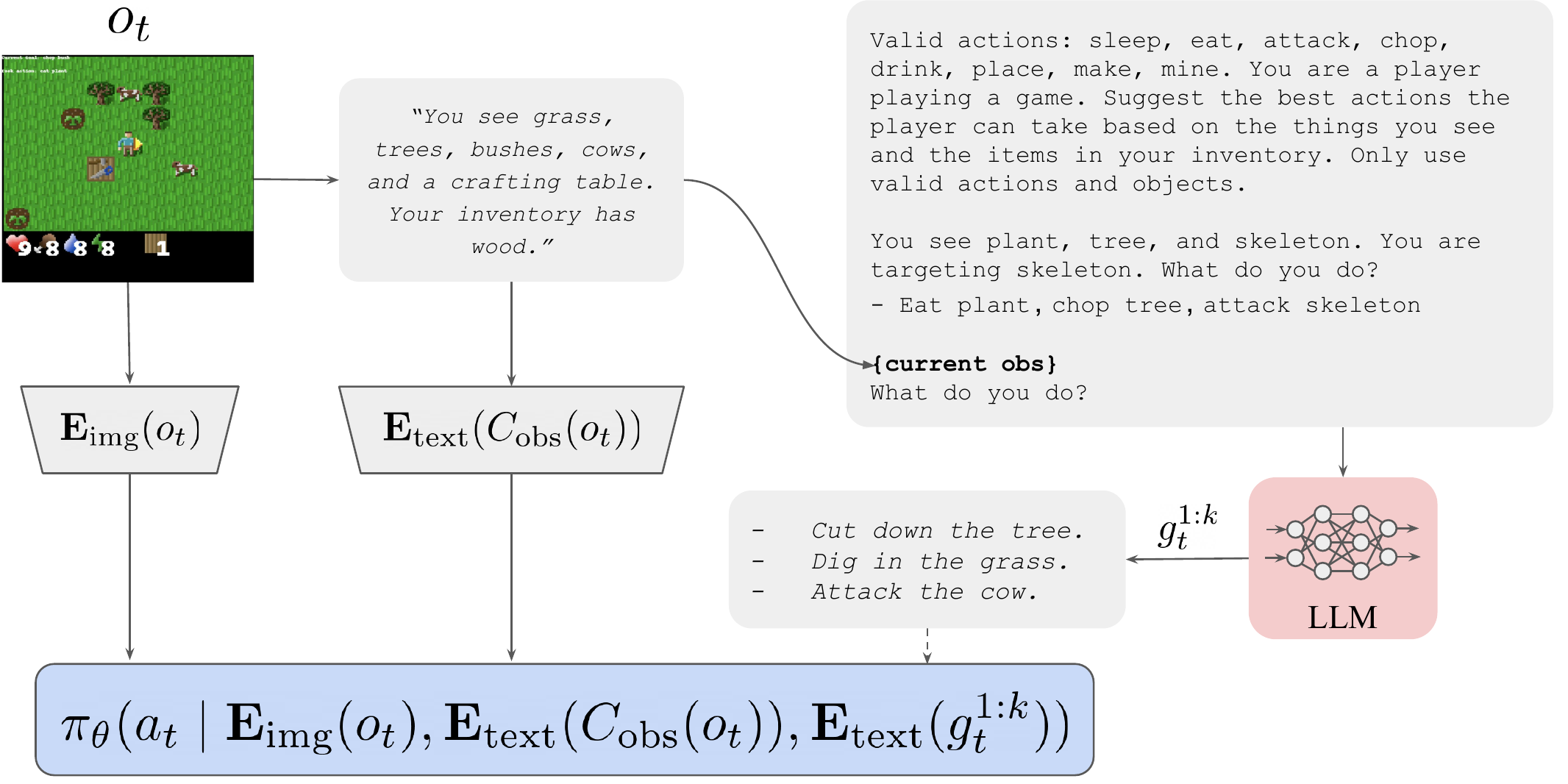} 
\caption{Policy parametrization for ELLM. We optionally condition on embeddings of the goals $E_\text{text}(g_t^{1:k})$ and state $E_\text{text}(C_\text{obs}(o_t))$.}
\end{subfigure}
\label{fig:policyparam}
\begin{subfigure}{.38\textwidth}
\centering
\includegraphics[width=\textwidth]{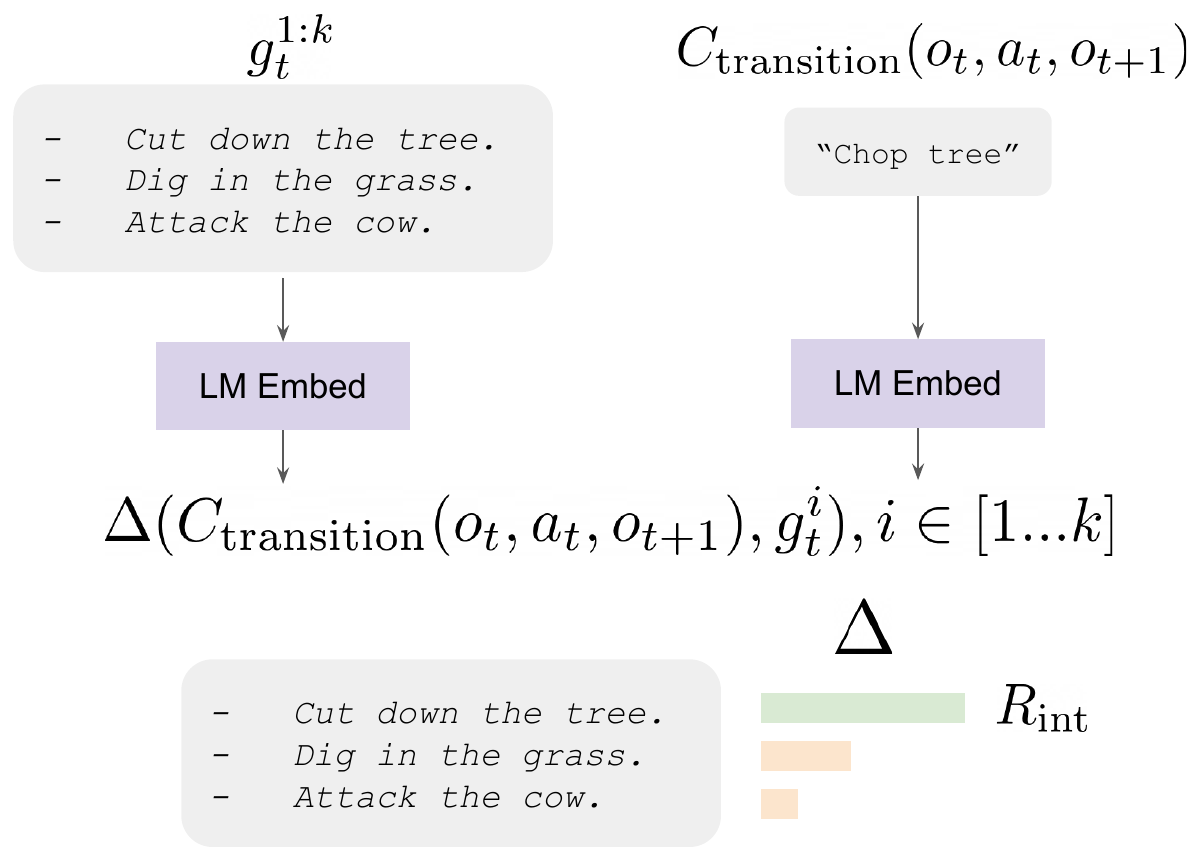} 
\caption{LLM reward scheme. We reward the agent for the similarity between the captioned transition and the goals.}
\end{subfigure}
\caption{ELLM uses GPT-3 to suggest adequate exploratory goals and SentenceBERT embeddings to compute the similarity between suggested goals and demonstrated behaviors as a form of intrinsically-motivated reward.}
\label{fig:expl_pipeline}
\end{figure*}

\section{Structuring Exploration with LLM Priors}\label{sec:method}

\subsection{Problem Description}\label{sec:method_ellm}

We consider partially observed Markov decision processes defined by a tuple $(\mathcal{S}, \mathcal{A}, \mathcal{O}, \Omega, \mathcal{T}, \gamma, \mathcal{R})$, in which observations $o\in\Omega$ derive from environment states $s\in\mathcal{S}$ and actions $a\in\mathcal{A}$  via $\mathcal{O}(o \mid s, a)$. $\mathcal{T}(s' \mid s, a)$ describes the dynamics of the environment while $\mathcal{R}$ and $\gamma$ are the environment's reward function and discount factor.

IM agents optimize for an intrinsic reward $\mathcal{R}_\text{int}$ alongside or in place of $\mathcal{R}$. CB-IM methods, in particular, define $\mathcal{R}_\text{int}$ via a family of goal-conditioned reward functions:
\begin{equation}
\label{eq:cbim}
\mathcal{R}_\text{int}(o,\,a,\,o’)\,=\,\mathbb{E}_{g\sim\red{\mathcal{G}}}\left[ \blue{\mathcal{R}_\text{int}(o,\,a,\,o’ \mid g)} \right].
\end{equation} 
 A CB-IM agent is expected to perform well with respect to the original $\mathcal{R}$ when the intrinsic reward $\mathcal{R}_\text{int}$ is both easier to optimize and well aligned with $\mathcal{R}$, such that behaviors maximizing $\mathcal{R}_\text{int}$ also maximize $\mathcal{R}$. Every CB-IM algorithm must define two elements in Equation~\ref{eq:cbim}: \red{(1)}~the distribution of goals to sample from, i.e. \red{$\mathcal{G}$}, and \blue{(2)}~the goal-conditioned reward functions \blue{$\mathcal{R}_\text{int}(o, a, o' \mid g)$}. Given these, A CB-IM algorithm trains a goal-conditioned policy $\pi(a \mid o, g)$ to maximize $R_\text{int}$. For some intrinsic reward functions,  agents may achieve high reward under the original reward function $\mathcal{R}$ immediately; for others, additional fine-tuning with $\mathcal{R}$ may be required.
In \cref{eq:cbim}, the space of goals $\mathcal{G}$ is determined by the goal-conditioned reward function $R_\text{int}(\cdot \mid g)$: every choice of $g$ induces a corresponding distribution over optimal behaviors. 

%

\subsection{Goal-based Exploration Desiderata}\label{sec:desiderata}
How should we choose \red{$\mathcal{G}$} and \blue{$\mathcal{R}_\text{int}(\cdot \mid g)$} to help agents make progress toward general reward functions $\mathcal{R}$?
Goals targeted during exploration should satisfy three properties:

\begin{itemize}
    \item  \textbf{Diverse}: targeting diverse goals increases the chance that the target behavior is similar to one of them.
    \item \textbf{Common-sense sensitive}: learning should focus on feasible goals (\texttt{chop a tree} $>$ \texttt{drink a tree}) which are likely under the distribution of goals humans care about (\texttt{drink water} $>$ \texttt{walk into lava}). 
    \item \textbf{Context sensitive}: learning should focus on goals that are feasible in the current environment configuration (\eg \texttt{chop a tree} only if a tree is in view).
\end{itemize}

Most CB-IM algorithms hand-define the reward functions \blue{$R_\text{int}$} \blue{(2)} and the support of the goal distribution \red{(1)} in alignment with the original task $\mathcal{R}$, but use various intrinsic motivations to guide goal sampling \red{(1)}: \eg novelty, learning progress, intermediate difficulty \citep[see a review in][]{colas2022autotelic}. In \textbf{E}xploring with \textbf{L}arge \textbf{L}anguage \textbf{M}odels (ELLM), we propose to leverage language-based goal representations and language-model-based goal generation to alleviate the need for environment-specific hand-coded definitions of \red{(1)} and \blue{(2)}. We hypothesize that world knowledge captured in LLMs will enable the automatic generation of goals that are diverse, human-meaningful and context sensitive.

\subsection{Goal Generation with LLMs $\mathbf{(\red{\mathcal{G}})}$}

Pretrained large language models broadly fall into three categories: autoregressive, masked, or encoder-decoder models \cite{min2021recent}. Autoregressive models (\eg GPT; \citealp{radford2018improving}), are trained to maximize the log-likelihood of the next word given all previous words, and are thus capable of language generation. Encoder-only models (\eg BERT; \citealp{devlin2018bert}), are trained with a masked objective, enabling effective encoding of sentence semantics. Pretraining LMs on large text corpora yields impressive zero- or few-shot 
on diverse language understanding and generation tasks, including tasks requiring not just linguistic knowledge but world knowledge \cite{https://doi.org/10.48550/arxiv.2005.14165}. 

ELLM uses autoregressive LMs to 
generate goals
and masked LMs to 
build vector representations of goals.
When LLMs generate goals, the support of the goal distribution becomes as large as the space of natural language strings. 
While querying LLMs unconditionally for goals can offer diversity and common-sense sensitivity, context-sensitivity requires knowledge of agent state. Thus, at each timestep we acquire goals by prompting the LLM with a list of the agent's available actions and a text description of the current observation via a \textit{state captioner} $C_{\text{obs}}: \Omega \rightarrow \Sigma^*$, where $\Sigma^*$ is the set of all strings (see Figure \ref{fig:expl_pipeline}).

We investigate two 
concrete strategies for extracting goals from LLMs:
(1)~open-ended generation, in which the LLM outputs text descriptions of suggested goals (\eg \texttt{next you should...}), and (2)~closed-form, in which a possible goal is given to the LLM as a QA task (\eg \texttt{Should the agent do X? (Yes/No)}). Here the LLM goal suggestion is only accepted when the log-probability of \texttt{Yes} is greater than \texttt{No}. The former is more suited for open-ended exploration and the latter is more suited for environments with large but delimitable goal spaces. \rebuttal{While the LLM does not have prior knowledge of all possible goals, we can provide some guidance towards desirable suggestions through few-shot prompting. See Appendix \ref{sec:crafter-prompt} for the full prompt.}

\subsection{Rewarding LLM Goals $\mathbf{(\blue{R_\text{int}})}$}
Next we consider the goal-conditioned reward \blue{(2)}.
We compute rewards for a given goal $g$ ($\mathcal{R}_\text{int}$ in Eq.~\ref{eq:cbim}) by measuring the semantic similarity between the LLM-generated goal and the description of the agent's transition in the environment as computed by a \textit{transition captioner} $C_{\text{transition}}: \Omega\times\mathcal{A}\times\Omega \rightarrow \Sigma$:
\begin{equation*}
 \mathcal{R}_\text{int}(o, a, o' \mid g) =
\begin{cases}
    \Delta( C_{\text{transition}}(o,a,o'), g)& \text{if } > T\\
    0              & \text{otherwise.}
\end{cases}
\end{equation*}
Here, the semantic similarity function $\Delta (\cdot\,, \,\cdot)$ is defined as the cosine similarity between \emph{representations} from an LM encoder $E(\cdot)$ of captions and goals: 
\begin{equation*}
    \Delta (C_{\text{transition}}(o,a,o'), g) = \frac{E(C_{\text{transition}}(o,a,o')) \cdot E(g)}{\|E(C_{\text{transition}}(o,a,o'))\| \|E(g)\|}.
\end{equation*}
\rebuttal{In practice, we use a pretrained SentenceBERT model \cite{reimers-2019-sentence-bert} for $E(\cdot)$. We choose cosine similarity to measure alignment between atomic agent actions and freeform LLM generations, as done in prior work \cite{huang2022language}.}
When the caption of a transition is sufficiently close to the goal description ($\Delta>T$), where $T$ is a similarity threshold hyperparameter, the agent is rewarded proportionally to their similarity. Finally, since there can be multiple goals suggested, we reward the agent for achieving any of the $k$ suggestions by taking the maximum of the goal-specific rewards:
\begin{equation*}
    \rebuttal{\Delta^\text{max} = \max_{i=1\ldots k}\Delta\left(C_{\text{transition}}(o_t,a_t,o_{t+1}), g_t^i\right).}
\end{equation*}

As a result, the general reward function of CB-IM methods from Equation~\ref{eq:cbim} can be rewritten:
\begin{equation}\label{eq:ellm}
    \mathcal{R}_{\text{int}}(o, a, o') = \mathbb{E}_{\textbf{LLM}(g^{1\,..\,k} | C_\text{obs}(o))} \left[\Delta^\text{max}\right].
\end{equation}

\subsection{Implementation Details}\label{sec:imp-details}
The full ELLM algorithm is summarized in Algorithm~\ref{alg:ellm}. See Figure~\ref{fig:suggestion_pipeline} for the high-level pipeline. To impose a novelty bias, we also filter out LM suggestions that the agent has already achieved earlier in the same episode. This prevents the agent from exploring the same goal repeatedly. \rebuttal{In Appendix \ref{sec:novelty-ablation} we show this step is essential to the method.}

 We consider two forms of agent training: (1)~a \textbf{goal-conditioned} setting where the agent is given a sentence embedding of the list of suggested goals, $\pi(a \mid o, E(g^{1:k}))$, and (2)~a \textbf{goal-free} setting where the agent does not have access to the suggested goals, $\pi(a \mid o)$. While $R_\text{int}$ remains the same in either case, training a goal-conditioned agent introduces both challenges and benefits: it can take time for the agent to learn the meaning of the different goals and connect it to the reward, but having a language-goal conditioned policy can be more amenable to downstream tasks than an agent just trained on an exploration reward. We also consider two types of \rebuttal{policy inputs}\---\,(1)~just the partially observed pixel observations, or (2)~the pixel observations combined with the embedded language-state captions $E(C_{\text{obs}}(o))$. Since (2) performs better (see analysis in Appendix \ref{sec:crafter_pretrain_all}), we use (2) for all paper experiments unless otherwise specified.
 All variants are trained with the DQN algorithm \citep{mnih2013playing}, with implementation details in Appendix \ref{sec:alg-details}.

This paper focuses on the benefits of LLM priors for RL exploration and mostly assumes a pre-existing captioning function. In simulation, this can be acquired for free with the ground truth simulator state. For real world applications, one can use object-detection \citep{zaidi2022survey}, captioning models \citep{stefanini2022show}, or action recognition models \citep{kong2022human}.  Alternatively, one could use multi-modal vision-language models with a similar LM component \citep{alayrac2022flamingo}. 
To test the robustness of our method under varying captioning quality, Section \ref{sec:captioner} studies a relaxation of these assumptions by looking at a variant of ELLM using a learned captioner trained on human descriptions.

\begin{algorithm}[t]
\caption{ELLM Algorithm}\label{alg:ellm}
\begin{algorithmic}
\State Initialize untrained policy $\pi$
\State $t$ $\leftarrow$  0
\State $o_t$ $\leftarrow$   \texttt{env.RESET()}
\While{$t <$ \texttt{max\_env\_steps}}
\State \textcolor{blue}{\# Generate $k$ suggestions, filtering achieved ones}
\State $g^{1:k}_t$ \textleftarrow \texttt{PREV\_ACHIEVED}(\texttt{LLM}($C_{\text{obs}}$($o_t$))) 
\State \textcolor{blue}{\# Interact with the environment}
\State $a_t \sim$ $\pi$($a_t | o_t$, E($C_{\text{obs}}$($o_t$))), E($g^{1:k}_t$)) 
\State $s_{t+1} \leftarrow$ \texttt{env.STEP}($a_t$) 
\State \textcolor{blue}{\# Compute suggestion achievement reward}
\State $r_t  \leftarrow 0$ 
\rebuttal{\State $\Delta^{max} \leftarrow \text{max}_{i=1 \ldots k} \Delta(C_{\text{transition}}(o_t, a_t, o_{t+1}), g^i_t)$}
\If{$\Delta^{max} >$ \texttt{threshold}}
        \State $r_t = \Delta^{max}$
    \EndIf
\State \textcolor{blue}{\# Update agent using any RL algorithm}
\State Buffer$_{t+1}$ \textleftarrow Buffer$_t \cup  (o_t, a_t, g^{1:k}_t, r_t, o_{t+1})$ 
\State $\pi$ \textleftarrow \texttt{UPDATE}($\pi$, Buffer$_{t+1}$)
\EndWhile
\end{algorithmic}
\end{algorithm}

\section{Experiments}

Our experiments test the following hypotheses: 
\begin{itemize}
\vspace{-.3cm}
    \item \textbf{(H1)}~Prompted pretrained LLMs can generate plausibly-useful exploratory goals satisfying the desiderata listed in Section \ref{sec:desiderata}: diversity, common-sense and context sensitivity. 
    \vspace{-.2cm}
    \item \textbf{(H2)}~Training an ELLM agent on these exploratory goals improves performance on downstream tasks compared to  methods that do not leverage LLM-priors. 
\end{itemize}
\vspace{-.2cm}

\begin{figure}[t!]
\includegraphics[width=\columnwidth]{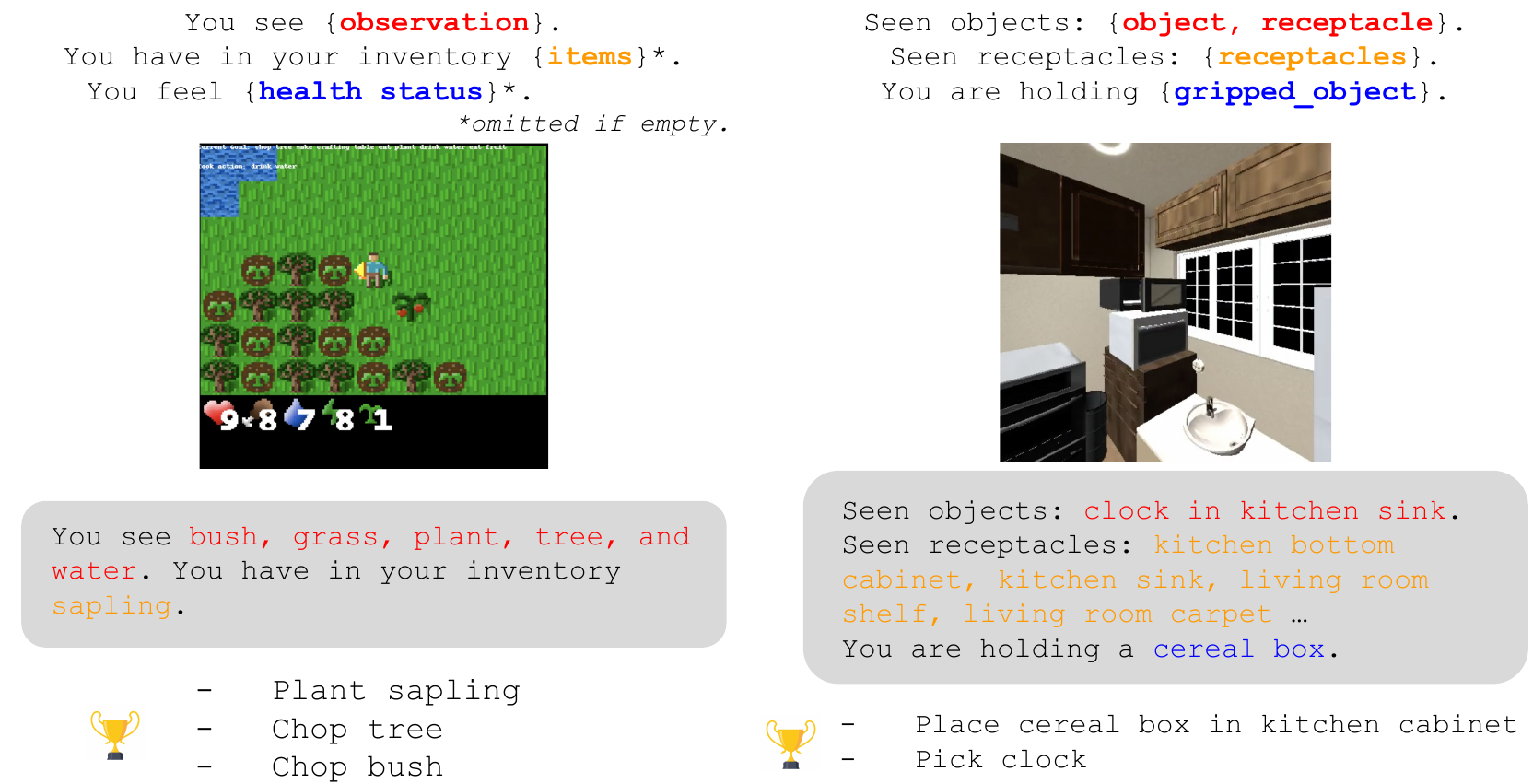}
\centering
\caption{Sample templated captions and \rebuttal{suggested goals}.}
\label{fig:example_caption}
\end{figure}

We evaluate ELLM in two complex environments: (1)~\textit{Crafter}, an open-ended environment in which exploration is required to discover long-term survival strategies \citep{hafner2021benchmarking}, and (2)~\textit{Housekeep}, an embodied robotics environment that requires common-sense to restrict the exploration of possible rearrangements of household objects \citep{housekeep}. Besides environment affordances, these environments also differ in viewpoint (3rd vs 1st person) and action space (large high-level vs low-level). In each environment, we compare ELLM with existing IM-RL methods \citep{liu2021behavior, burda2019exploration}, an oracle with ground-truth rewards, and ablations of ELLM; see Table~\ref{tab:baselines}.

\begin{table*}[]
\centering
\footnotesize
\renewcommand{\tabularxcolumn}{m} 
\begin{tabularx}{\textwidth}{c>{\raggedright}X}
 \toprule
\textbf{Method} & \textbf{Description} \tabularnewline
\midrule
ELLM (ours) & Rewards the agent for achieving any goal suggested by the LLM using the similarity-based reward functions $R_\text{int}$ defined in Eq.~\ref{eq:ellm}. It only rewards the agent for achieving a given goal once per episode (novelty bias).
\tabularnewline

\midrule
\textit{Oracle} & The upper bound: it suggests all context-sensitive goals at any step, only common-sensical ones (from the list of
\tabularnewline
(Crafter only) & valid goals) and uses the same novelty bias as ELLM. Rewards are computed exactly with a hard-coded $R_\text{int}$.
\tabularnewline

\midrule
Novelty & This baseline removes the common-sense sensitivity assumption of the \textit{Oracle} and rewards the agent for achieving any of the goals expressible in the environment including invalid ones (\eg \texttt{drink tree}) as long as the agent performs the goal-reaching action in the right context (\eg while facing a tree). Uses a hard-coded $R_\text{int}$ and a novelty bias like the \textit{Oracle}.
\tabularnewline

\midrule
Uniform & This variant removes the novelty bias from \textit{Novelty} and samples uniformly from the set of expressible goals. 
\tabularnewline

\midrule
APT  & State-of-the-art KB-IM algorithm that maximizes state entropy computed as the distance between the current 
\tabularnewline
{\scriptsize\citep{liu2021behavior}} &  state's embedding $e_s$ and its K nearest neighbors $e_{s^{[1..K]}}$ within a minibatch uniformly sampled from memory. There is no goal involved and $R_\text{int}=\log \|e_s - e_{s^{[1..K]}} \|$.
\tabularnewline
\midrule
RND  & State-of-the-art KB-IM algorithm that rewards the agent for maximizing a form of novelty estimated by the
\tabularnewline
{\scriptsize\citep{burda2019exploration}} & prediction error of a model $h$ trained to predict the output of a random network $\tilde{h}$. $R_\text{int}\,=\,\|h(s, a)\,-\,\tilde{h}(s, a)\|.$
\tabularnewline
\bottomrule
\end{tabularx}
\caption{Descriptions of the compared algorithms. (Additional comparisons in Appendix \ref{sec:noveld}).}
\label{tab:baselines}
\end{table*}

\subsection{Crafter} \label{sec:crafter}

\begin{table}[]
    \centering
    \footnotesize
    \begin{tabular}{lcc}
    \toprule
    & Suggested & Rewarded  \\
    \midrule
       Context-Insensitive & 13.6\% &  1.1\%  \\ 
       Common-Sense Insensitive & 16.4\%  & 32.4\%   \\   
       Good & 64.9\%  & 66.5\%   \\ 
       Impossible & 5.0\% & 0\%  \\
       \bottomrule
    \end{tabular}
    \caption{Fractions of suggested and rewarded goals that fail to satisfy context-sensitivity or common-sense sensitivity; that satisfy these properties and are achievable in Crafter (Good); or that are not allowed by Crafter's physics. See Appendix \ref{sec:crafter_llm} for examples of each.}
    \label{tab:crafter_lm_breakdown}
\end{table}

\paragraph{Environment description.} We first test ELLM in the Crafter environment, a 2D version of Minecraft \citep{hafner2021benchmarking}. Like Minecraft, Crafter is a procedurally generated and partially observable world that enables collecting and creating a set of artifacts organized along an achievement tree which lists all possible achievements and their respective prerequisites (see Figure~4 in \citealp{hafner2021benchmarking}). Although Crafter does not come with a single main task to solve, we can track agent progress along the achievement tree. 

We modify the original game in two ways. 
Crafter's original action space already incorporates a great deal of human domain knowledge: a single \texttt{do} action is interpreted in different ways based on the agent's context, each of which would correspond to a very different low-level action in a real environment (`\texttt{do}' means `\texttt{attack}' in front of a zombie but `\texttt{eat}' in front of a plant). 
We remove this assistance by augmenting the action space with more specific verb + noun pairs that are not guaranteed to be useful (\eg `\texttt{eat zombie}'). This makes it possible in Crafter to attempt a wide range of irrelevant/nonsensical tasks, providing an opportunity for an LM narrow the goal space down to reasonable goals.  See Appendix \ref{sec:crafter_action_space} for details. Second, to make RL training easier across all conditions, we increase the damage the agent does against enemies and reduce the amount of wood required to craft a table from 2 to 1; see Appendix Figure~\ref{fig:crafter_original} for comparisons. 

We use Codex \citep{chen2021evaluating} as our LLM with the open-ended suggestion generation variant of ELLM, where we directly take the generated text from the LLM as the set of suggested goals to reward.  Each query prompt consists of a list of possible verbs the agent can use (but not a list of all possible nouns), a description of the agent's current state, and the question `\texttt{What do you do?}'. We add two examples of similar queries to the start of the prompt in order to guide the language model to format suggestions in a consistent way; see the full prompt in Appendix \ref{sec:crafter-prompt}. 

\paragraph{Goals suggested by the LLM.}
To answer \textbf{H1}, we study the goals suggested by the LLM in Table~\ref{tab:crafter_lm_breakdown}: are they diverse, context-sensitive and common-sensical? The majority of suggested goals (64.9\%) are context-sensitive, sensible, and achievable in the game. Most of the 5\% of goals not allowed by Crafter's physics (\eg \texttt{build a house}) are context- and common-sensitive as well. The last third of the goals violate either context-sensitivity (13.6\%) or common-sense (16.4\%). See Appendix \ref{sec:crafter_llm} for details.

\paragraph{Pretraining exploration performance.}

\begin{figure}[]
\centering
\includegraphics[width=.9\columnwidth]{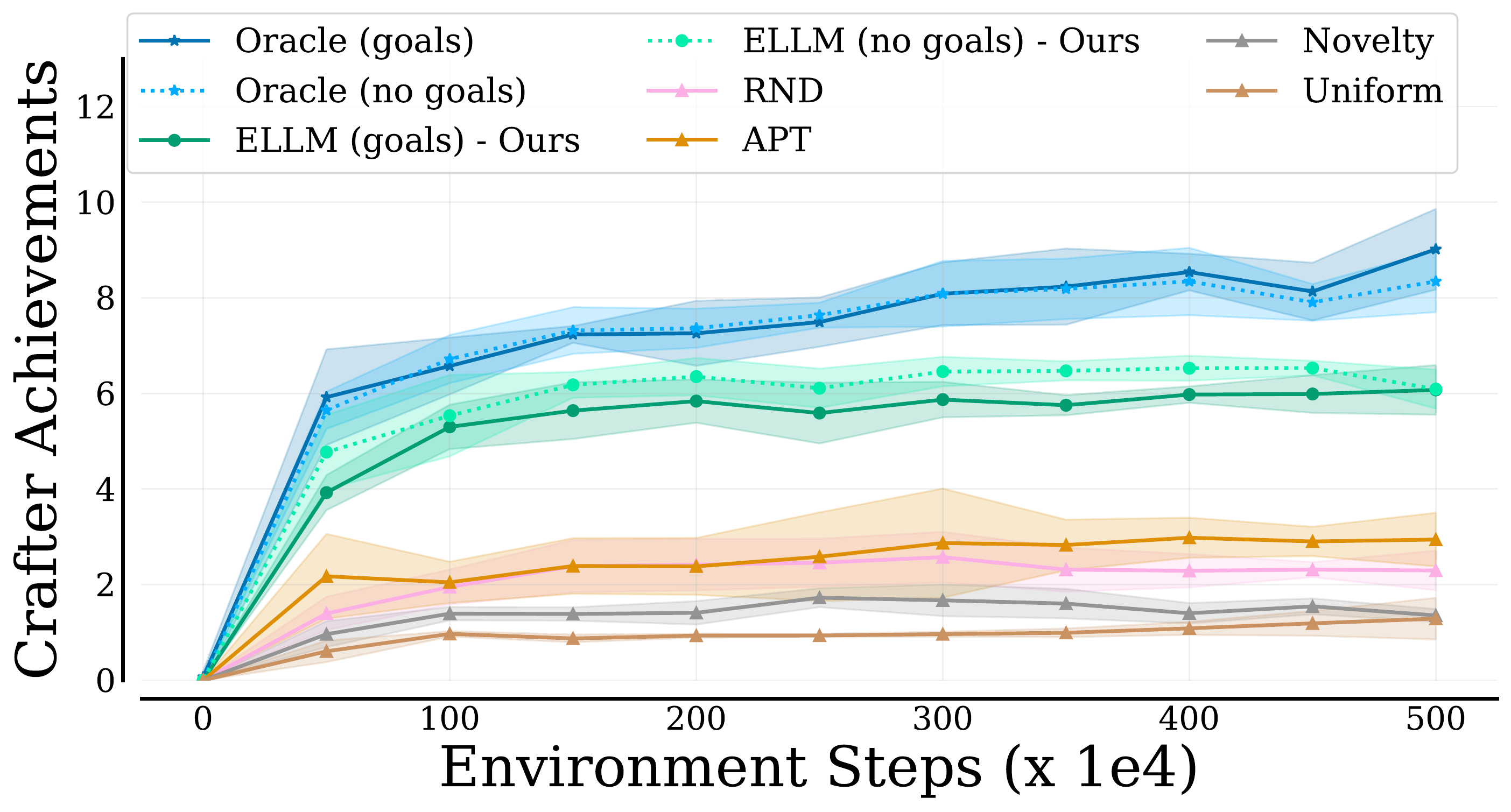}
\caption{Ground truth achievements unlocked per episode across pretraining, mean$\pm$std across 5 seeds.}
\label{fig:cpretraininga}
\end{figure}

A perfect exploration method would unlock all Crafter achievements in every episode, even without prior knowledge of the set of possible achievements. Thus, we measure exploration quality as the average number of unique achievements per episode across pretraining (Figure~\ref{fig:cpretraininga}). Although it is not given access to Crafter's achievement tree, ELLM learns to unlock about 6 achievements every episode, against 9 for the ground-truth-reward Oracle (Figure~\ref{fig:cpretraininga}). It outperforms all exploration methods that only focus on generating novel behaviors (APT, RND, Novelty)\,---\,all limited to less than 3 achievements in average.  As shown in Table~\ref{tab:crafter_lm_breakdown}, ELLM does not only focus on novelty but also generates common-sensical goals. This boosts exploration in Crafter, supporting \textbf{H1}.

As discussed in Section \ref{sec:imp-details}, we also test variants of each method (with / without goal conditioning, with / without text observations) where applicable.  We do not find goal conditioning to bring a significant advantage in performance during pretraining. The non-conditioned agent might infer the goals (and thus the rewarded behaviors) from context alone.  Similarly to \citet{mu2022improving} and \citet{ tam2022semantic}, we find that agents trained on visual + textual observations (as computed by $E(C_{\text{obs}}(o))$) outperform agents trained on visual observations only for all the tested variants (opaque vs semi-transparent bars in Appendix Figure~\ref{fig:cpretrainingb}). That said, optimizing for novelty alone, whether in visual or semantic spaces, seems to be insufficient to fully solve Crafter.

The na\"{i}ve approach of finetuning a pretrained policy on the downstream task performs poorly across all pretraining algorithms. We hypothesize this is because relevant features and Q-values change significantly between pretraining and finetuning, especially when the density of rewards changes. \rebuttal{Instead, we find it is more effective to use the pretrained policy for guided exploration. We initialize and train a new agent, but replace 50\% of the algorithm's randomly-sampled $\epsilon$-greedy exploration actions with actions sampled from the pretrained policy.  In Appendix \ref{sec:expl_vs_finetune} we include the poor finetuning results discuss why we think guided exploration does better. }

\begin{figure*}[ht!]
\includegraphics[width=\textwidth]{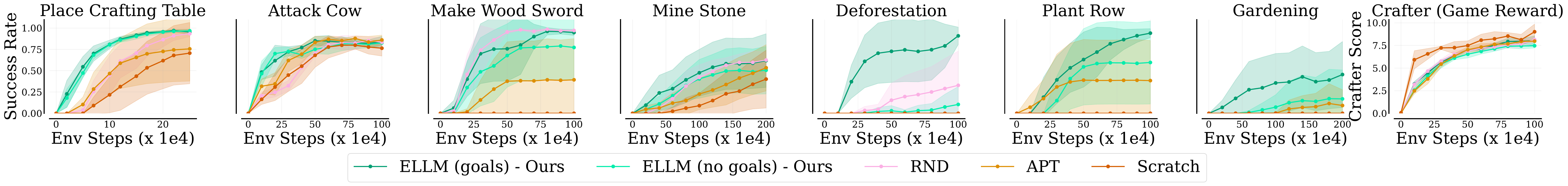}
\caption{Success rates across training for each of the seven downstream tasks in the Crafter environment. Each run trains an agent from scratch while leveraging a pretrained policy for exploration. Plots show mean $\pm$ std for 5 seeds. Some plots have multiple overlapping curves at 0.}
\label{fig:crafter_finetune}
\end{figure*}

Figure~\ref{fig:crafter_finetune} compares the downstream performance of ELLM to the performance of the two strongest baselines RND and APT using both transfer methods. (full comparisons with all baselines shown in Appendix \ref{sec:crafter_finetune_full}). For the goal-conditioned version of ELLM, we provide the agent with the sequence of subgoals required to achieve the task. Even though not all subgoals were mastered during pretraining, we still observe that the goal-conditioned pretrained agents outperform the unconditioned ones.

Performance of the different methods varies widely task-to-task and even seed-to-seed since each task requires a different set of skills, and any given agent may or may not have learned a particular skill during pretraining. \rebuttal{For instance, ELLM agents typically learn to place crafting tables and attack cows during pretraining, leading to low-variance learning curves. They typically do not learn to make wood swords, so we see a high-variance learning curve which depends on how quickly each agent stumbles across the goal during finetuning.} Despite the variance, we see that goal-conditioned ELLM stands out as the best-performing method on average. Notably, ELLM (both goal-conditioned and goal-free) is the only method with nonzero performance across all tasks.

\paragraph{ELLM with imperfect transition captioner.} \label{sec:captioner}

\begin{figure}[t!]
\centering
\includegraphics[width=.9\columnwidth]{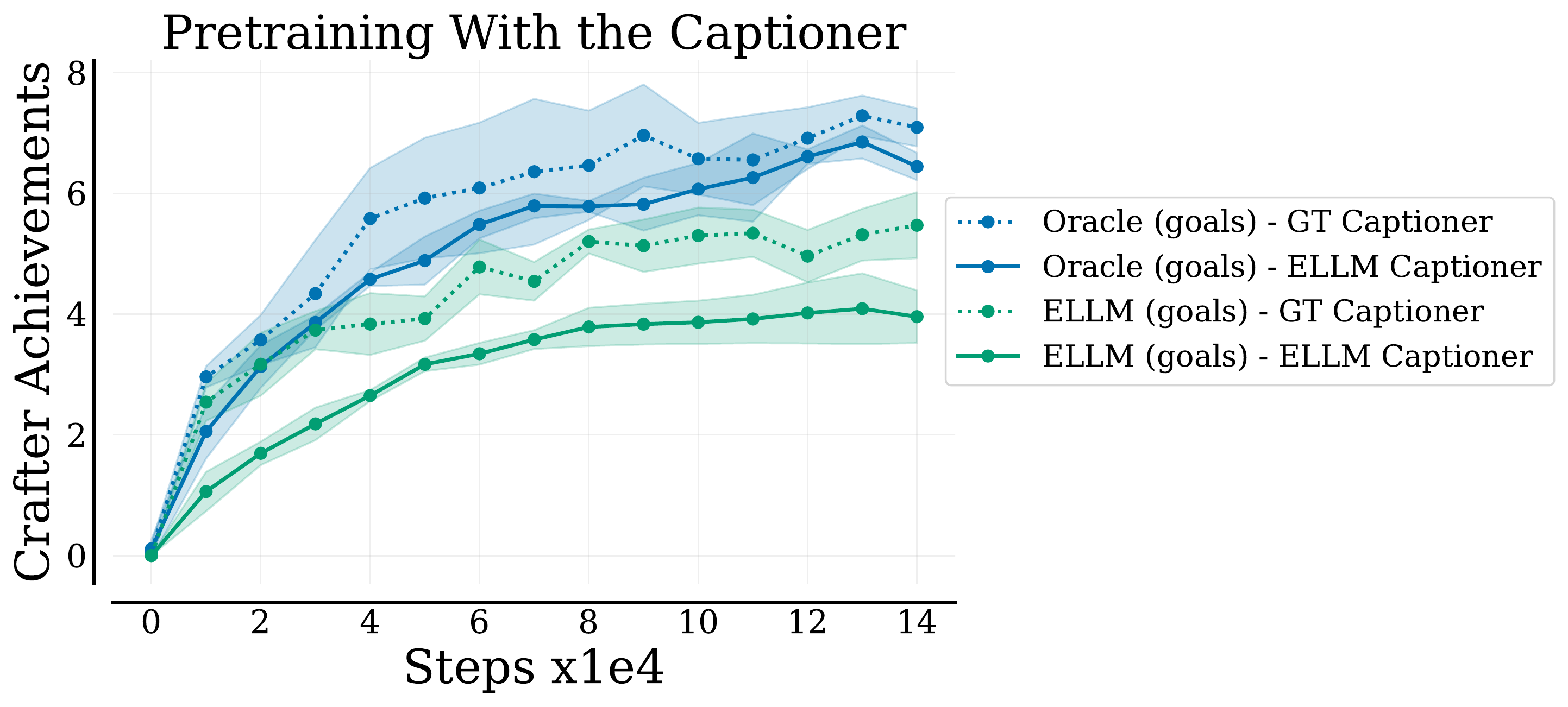}
\caption{Pretraining with a learned captioner vs a ground truth captioner. We see performance drops, especially for ELLM, but still relatively good performance. (3 seeds, mean$\pm$ std.)}
\label{fig:captioner}
\end{figure}

Perfect captioners might not be easy to obtain in some environments. However, trained captioners might generate more linguistic diversity and make mistakes. To test the robustness of ELLM to diverse and imperfect captions, we replace the oracle transition captioner $C_{\text{transition}}$ with a captioner trained on a mixture of human and synthetic data (847+900 labels) using the ClipCap algorithm \citep{mokady2021clipcap}. Synthetic data removes some of the human labor while still providing a diversity of captions for any single transition (3 to 8). Appendix~\ref{sec:captioner_appendix} presents implementation details and analyzes how the trained captioner might cause errors in generated rewards. Although its false negative rate is low (it detects goal achievements well), its false positive rate is rather high. This means it might generate rewards for achievements that were not unlocked due to a high similarity between the generated caption and goal description generated by the LLM. In ELLM pretraining, we use the learned captioner to caption transitions where an action is successful and use that caption to compute the reward via the similarity metric (see Section~\ref{sec:method}). Figure~\ref{fig:captioner} shows that ELLM performance is overall robust to this imperfect captioner.

\begin{figure*}[ht!]
\centering
\begin{subfigure}{\linewidth}
  \centering
\includegraphics[width=\columnwidth]{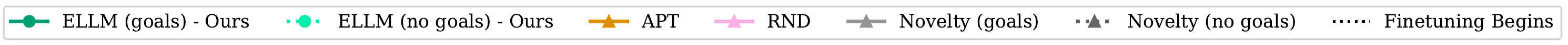}
  \end{subfigure} 
\begin{subfigure}{.49\linewidth}
\begin{subfigure}{.26\textwidth}
\includegraphics[width=\columnwidth]{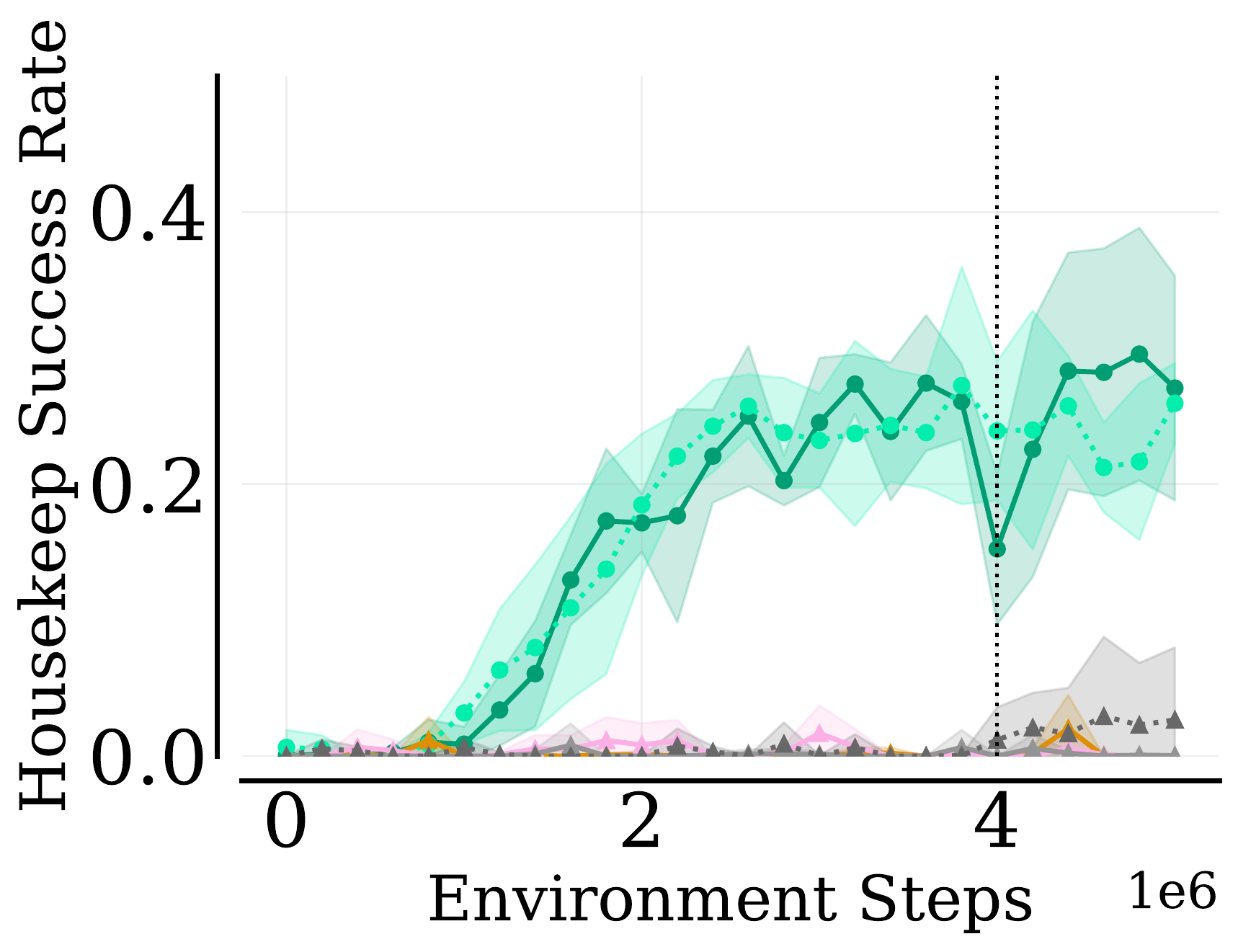}
\end{subfigure}
\begin{subfigure}{.23\textwidth}
\includegraphics[width=\columnwidth]{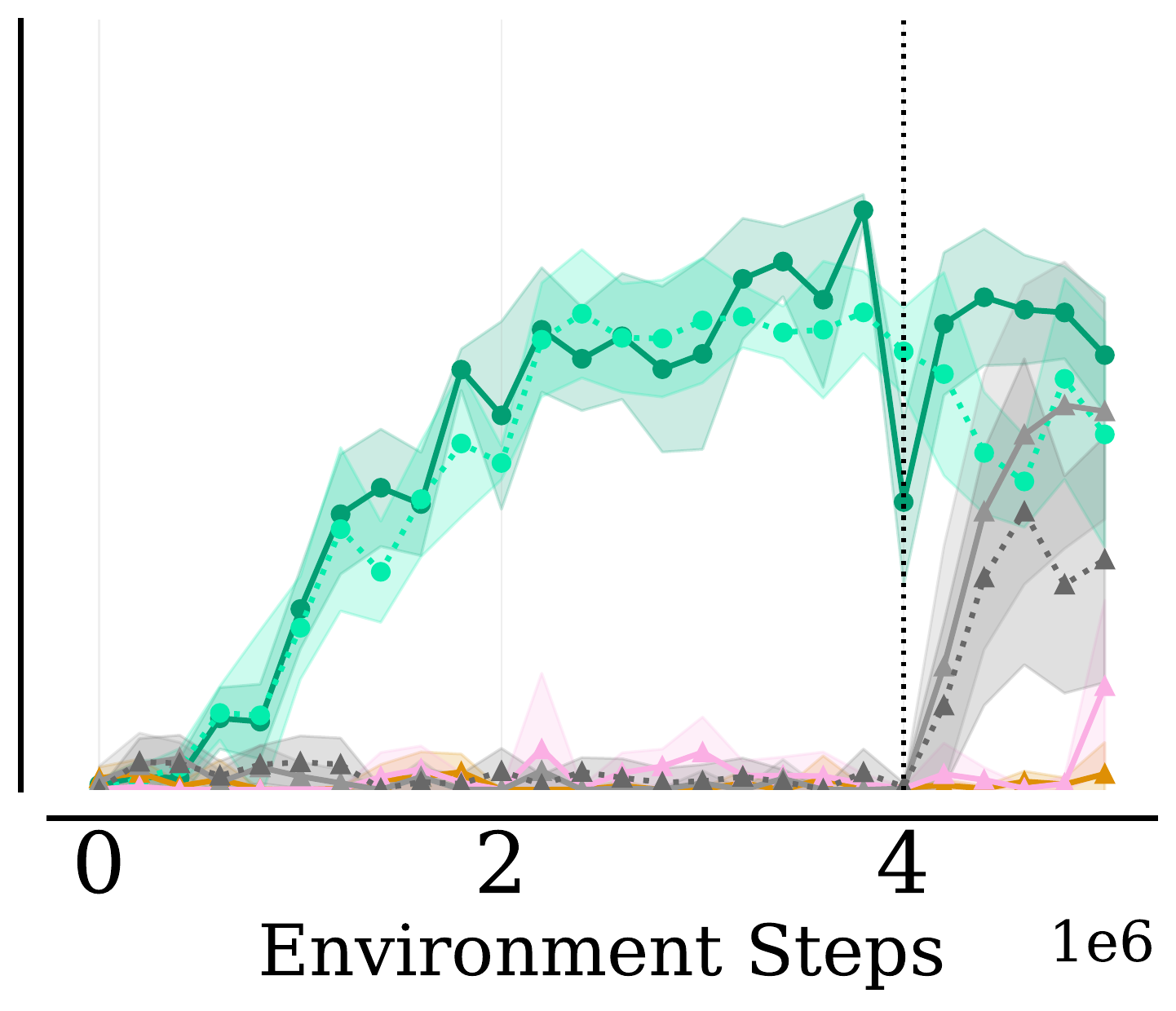}
\end{subfigure}
\begin{subfigure}{.23\textwidth}
\includegraphics[width=\columnwidth]{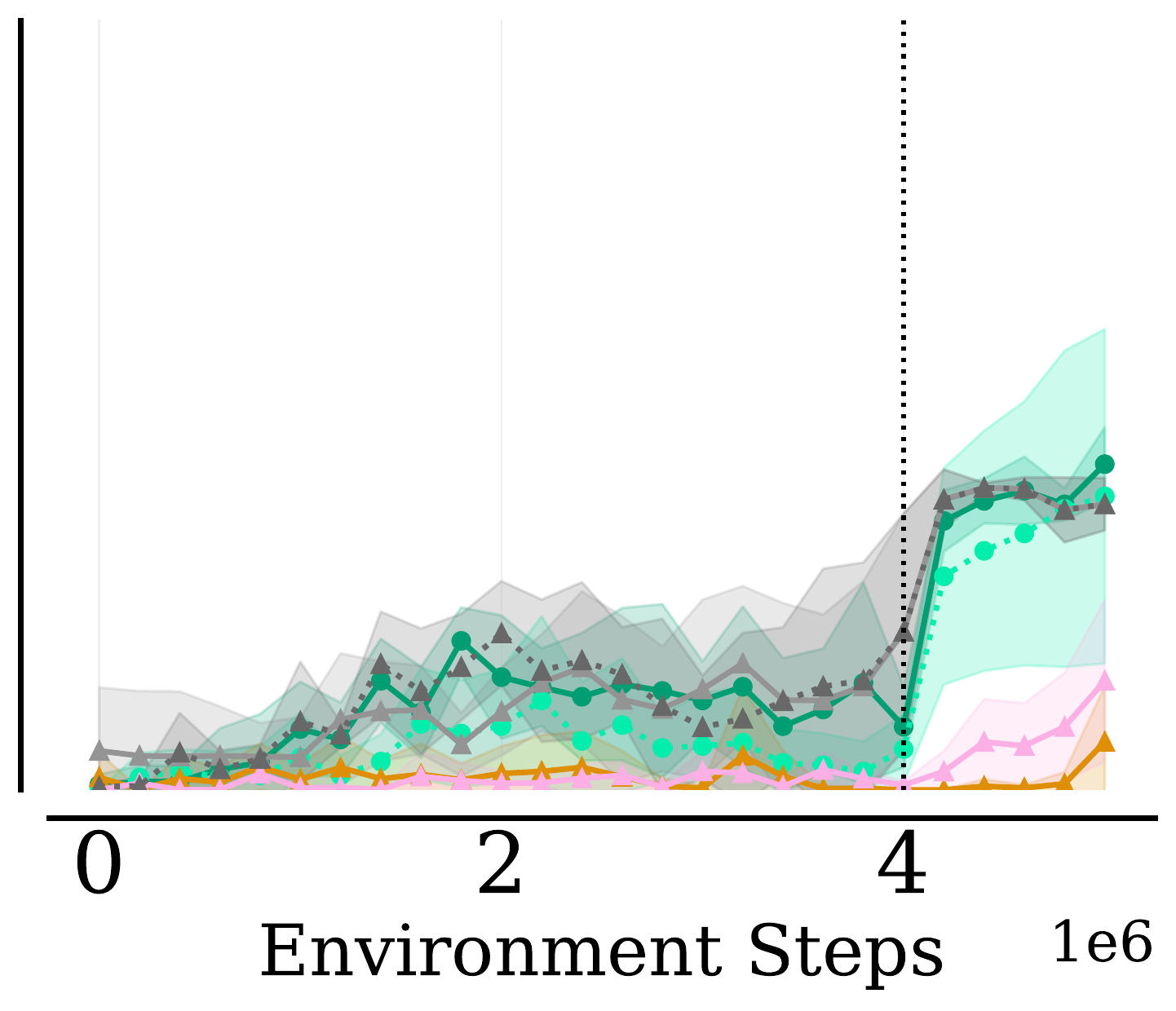}
\end{subfigure}
\begin{subfigure}{.23\textwidth}
\includegraphics[width=\columnwidth]{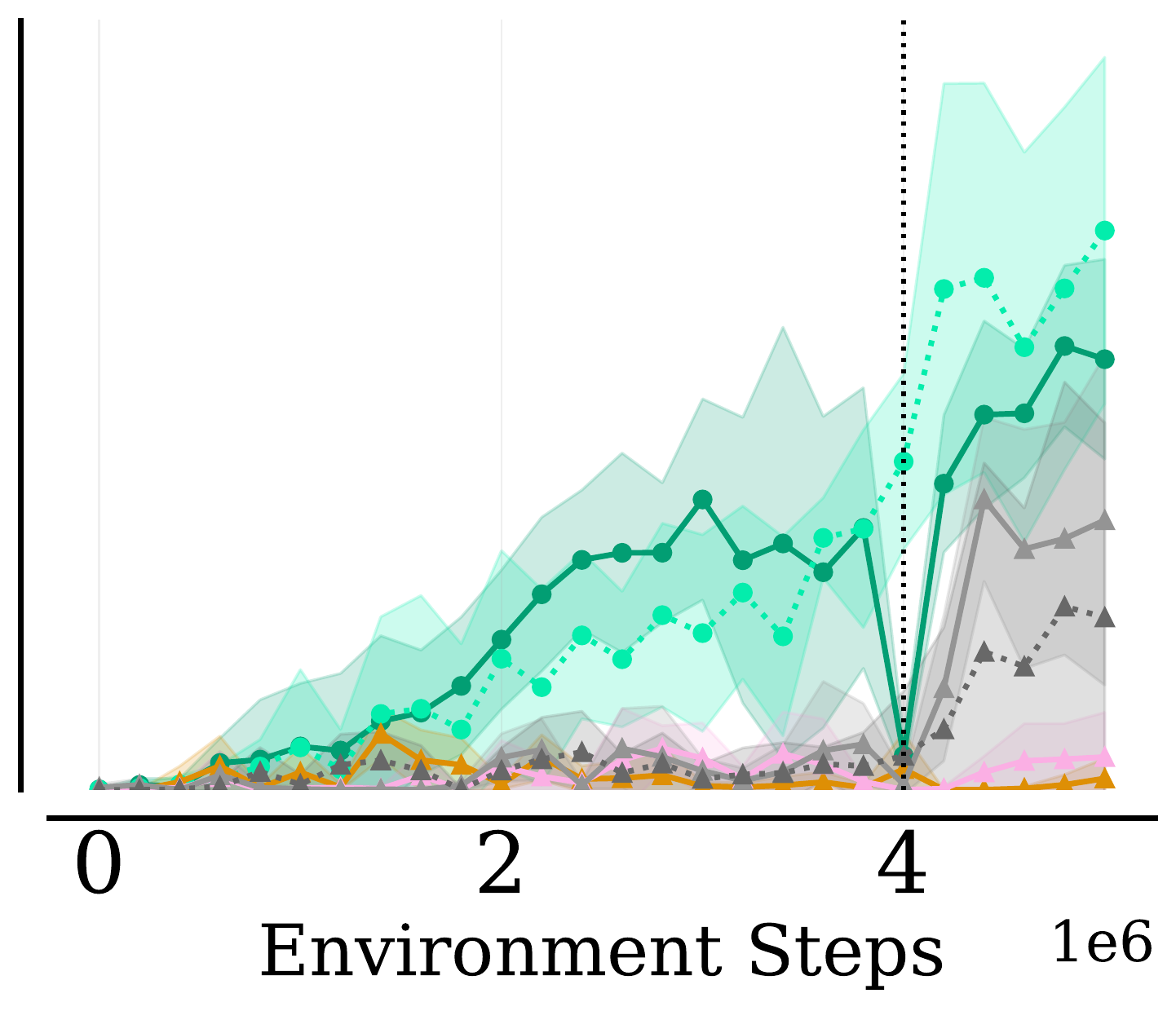}
\end{subfigure}
\caption{ \textit{Pretraining and finetuning:} pretraining for 4M steps then finetuning for 1M steps on the ground truth correct arrangement.}
\label{fig:housekeep-ptft}
\end{subfigure}
\centering
\begin{subfigure}{.5\linewidth}
\begin{subfigure}{.26\textwidth}
\includegraphics[width=\columnwidth]{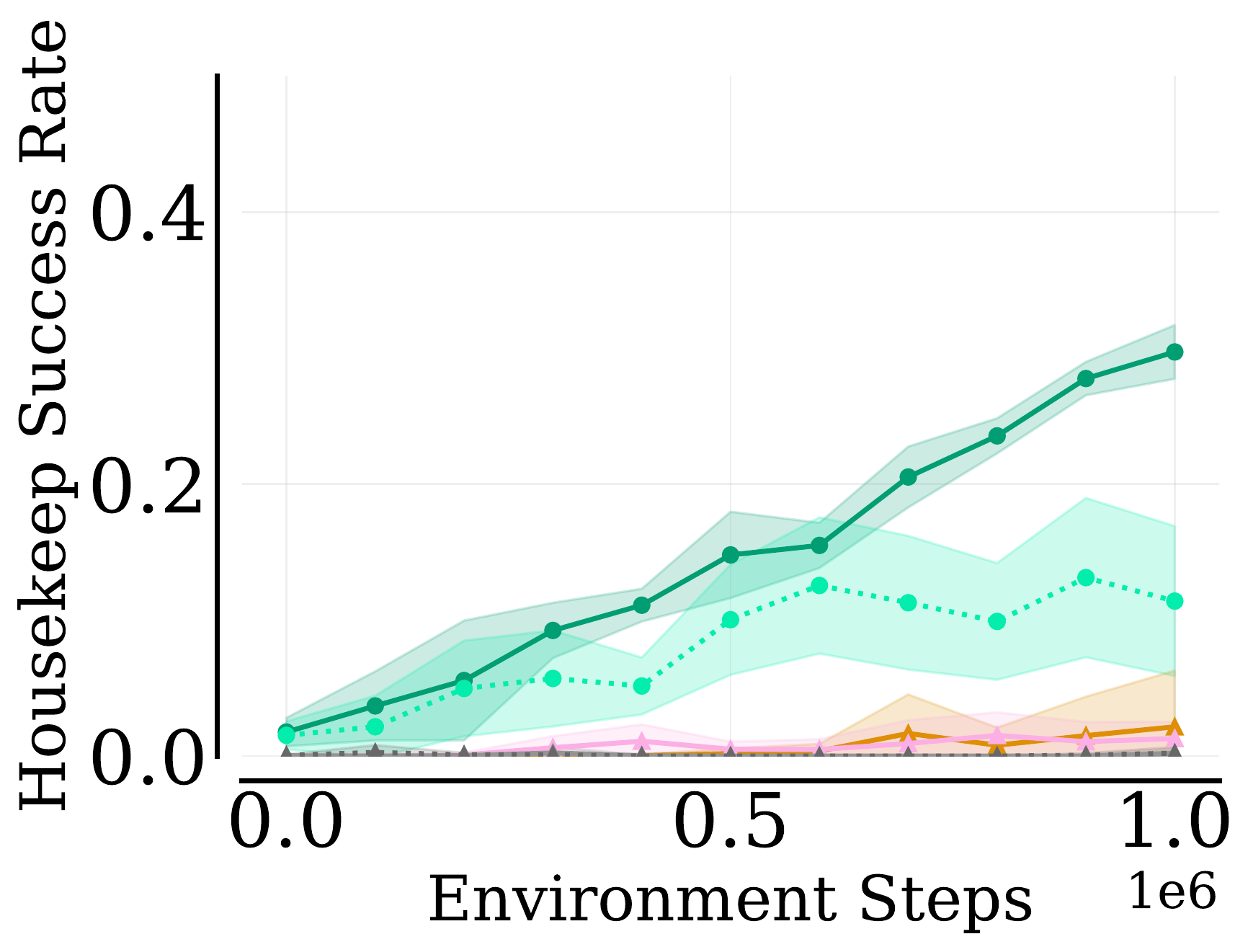}
\end{subfigure}
\begin{subfigure}{.23\textwidth}
\includegraphics[width=\columnwidth]{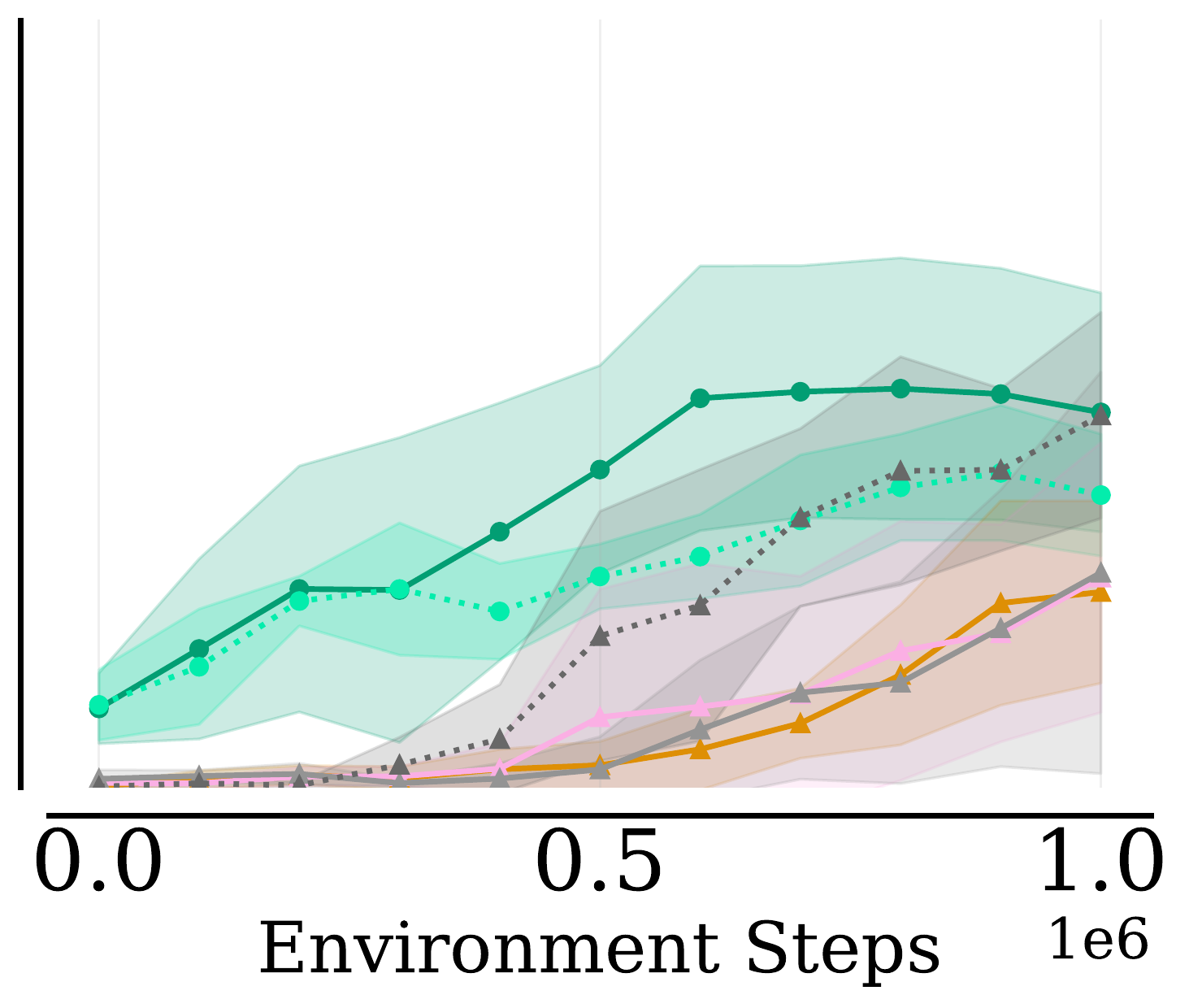}
\end{subfigure}
\begin{subfigure}{.23\textwidth}
\includegraphics[width=\columnwidth]{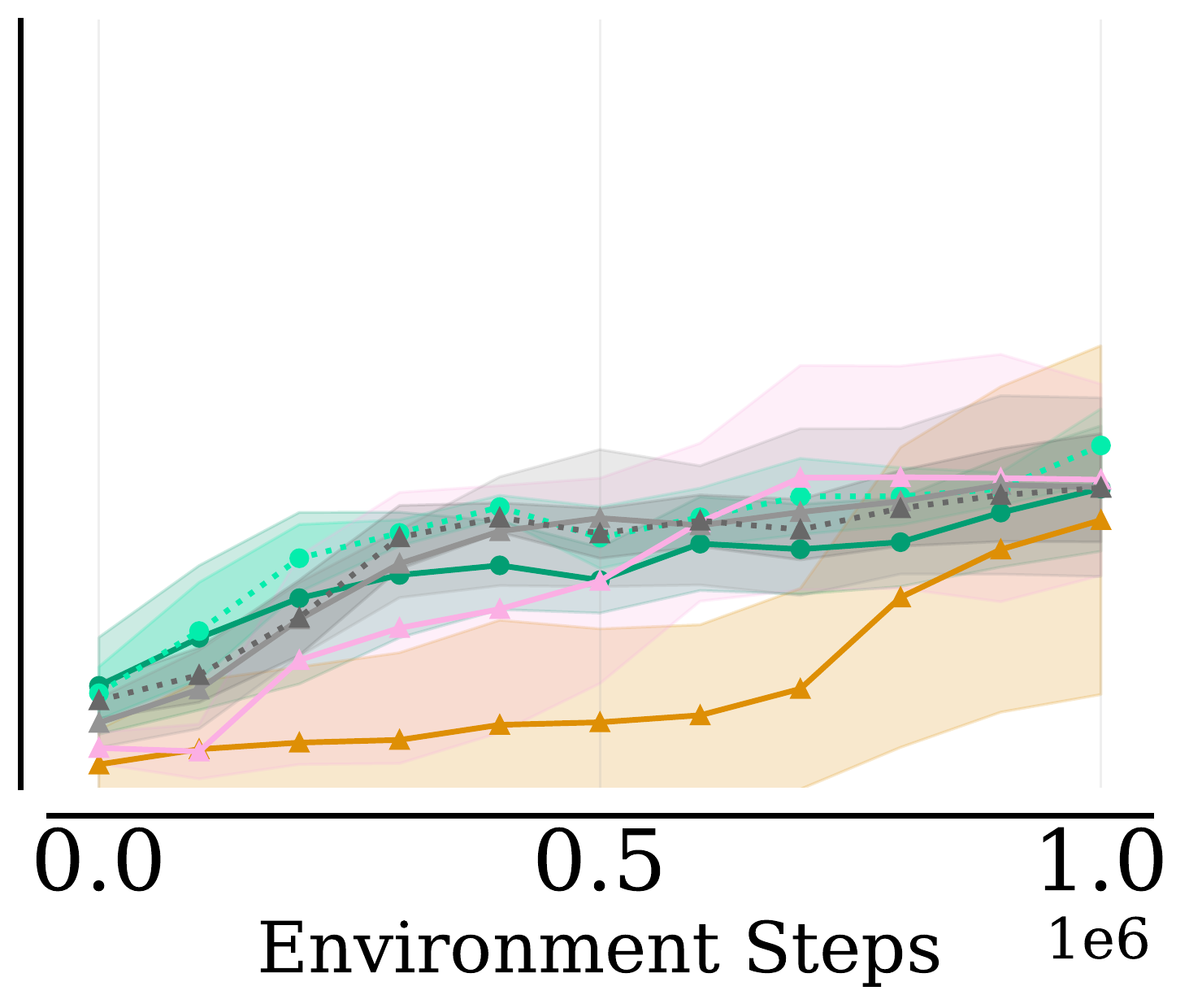}
\end{subfigure}
\begin{subfigure}{.23\textwidth}
\includegraphics[width=\columnwidth]{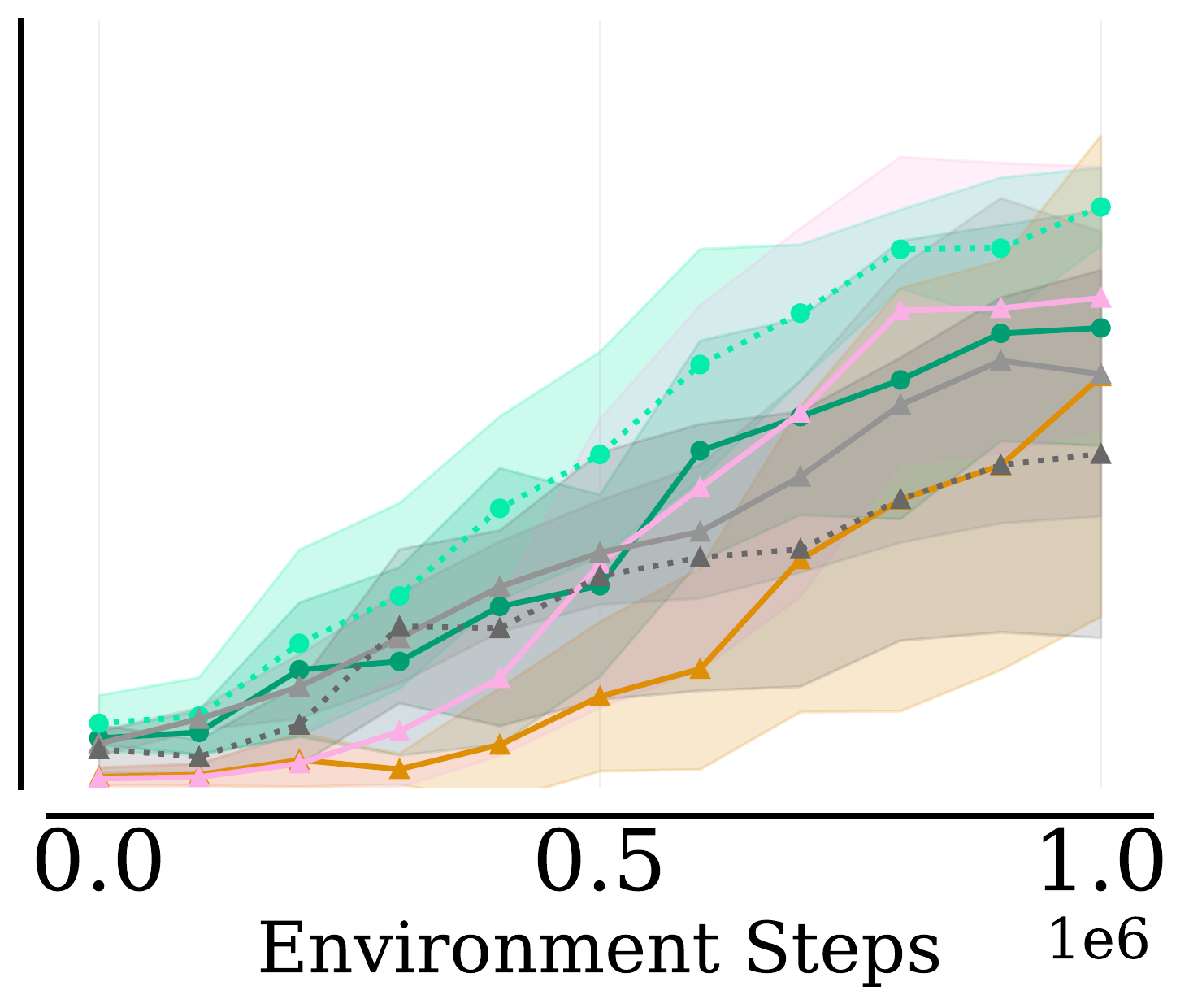}
\end{subfigure}
\caption{\textit{Downstream evaluation: }Using the frozen pretrained exploration policies only for $\epsilon$-greedy-style action selection for 1M steps.}
\label{fig:housekeep-ft}
\end{subfigure}
\caption{\textit{Housekeep:} Correct arrangement success rates on 4 object-receptacle task sets. Mean $\pm$ std over 5 seeds. }
\end{figure*} 

\subsection{Housekeep}

\textbf{Environment description.} Housekeep is an embodied robotics environment where the agent is tasked with cleaning up a house by rearranging misplaced objects \citep{housekeep}. The agent must successfully match the environment's ground truth correct mapping of objects to receptacles without direct instructions specifying how objects need to be rearranged. This mapping was determined via crowd-sourcing common-sense object-receptacle combinations. An example layout of the task can be found in Figure~1 in \citet{housekeep}. Common-sense priors are necessary for learning to rearrange misplaced objects into reasonable configurations.

We focus on a simplified subset of Housekeep consisting of 4 different scenes with one room each, each with 5 different misplaced objects and a suite of different possible receptacles; see Appendix~\ref{sec:housekeep-tasks} for details. Because the agent does not have access to the ground truth target locations, we use the game reward's rearrangement success rate as a measure of exploration quality: common-sensical exploration should perform better. A success rate of 100\% means the agent has picked and placed all 5 misplaced objects in correct locations. Note that we intentionally focus on a domain where the downstream application benefits strongly from exploring reasonable goals during pretraining. Rather than designing reward functions that correspond to all correct rearrangements for all possible objects, we investigate whether ELLM can be a general purpose method that guides learning human-meaningful behaviors.

Unlike Crafter's combinatorial and high-level action space, Housekeep operates with low-level actions: moving forward, turning, looking up or down, and picking or placing an object. This allows us to investigate whether ELLM enables high-level exploration despite using lower-level control.  We assume access to an egocentric instance segmentation sensor to generate captions of in-view objects and receptacles, and use the \texttt{text-davinci-002} InstructGPT model \cite{ouyang2022training} as our LLM. Given a description of visible objects, the receptacles the objects are currently in, and all previously seen receptacles, we create a list of all possible object-receptacle mappings. We use the closed-form variant of ELLM and query the LLM for whether each object should be placed in each receptacle as a \texttt{yes/no} question. By querying for each object-receptacle combination individually, we are able to cache and efficiently reuse LLM queries. The agent can be given two types of goals: (1) picking an object if it is not already in a suggested receptacle, and (2) placing a gripped object in a suggested receptacle.

\paragraph{Goals suggested by LLM.}
\begin{table}[t!]
    \centering
    
    \footnotesize
    \begin{tabular}{lcccc}
    \toprule
    & \textbf{Task 1} & \textbf{Task 2} & \textbf{Task 3} & \textbf{Task 4}  \\
    \midrule
       \textbf{Match Acc.} &  85.7\% & 87.5\% & 50\% & 66.7\% \\ 
       \textbf{Mismatch Acc.} & 93.8\% & 90.1\% & 94.0\% & 87.6\% \\ 
       \bottomrule
    \end{tabular}
    \caption{Classification accuracy of LLM for each Housekeep task (top row is true positives, bottom row is true negatives).}
    \label{tab:housekeep-llmacc}
\end{table}
In Housekeep, we assess LLM goals by looking at the classification accuracy of correct and incorrect arrangements (Table~\ref{tab:housekeep-llmacc}). We find that the LLM accuracy at identifying mismatches (\eg \texttt{vase in kitchen sink}) are all above 87\%, however, accuracy of identifying matches varies greatly depending on the available objects and receptacles (ranging from 50-90\%). Since there are only a few correct positions, each false negative hurts accuracy greatly. Taking a closer look, we find that some LLM labels are reasonable despite disagreeing with the environment mapping: \eg suggesting \texttt{vase in living room table}, and not suggesting \texttt{pan in living room cabinet}. This suggests that there are ambiguities in the ground truth mappings, likely due to human disagreement.

\paragraph{Pretraining and downstream performance.}

To investigate \textbf{H1}, we compare ELLM against the strongest baselines (RND, APT, Novelty) described in Table~\ref{tab:baselines}. In Housekeep the novelty baseline rewards the agent for novel instances of pick or place actions in an episode, allowing us to differentiate between success attributable solely to the captioner and the pick/place prior, and success attributable to any LLM common-sense priors. For brevity, we focus only on the pixel + text-observation variant of all methods.  Sample efficiency curves measuring the ground truth rearrangement success during both pretraining and finetuning are shown in Figure~\ref{fig:housekeep-ptft}. In three of the four tasks, we find that the ELLM bias leads to higher success rates during pretraining, suggesting coverage better aligned with the downstream task compared to the baselines.  We also find much higher pretraining success rates in the first two tasks. Since Table~\ref{tab:housekeep-llmacc} shows higher LLM accuracy for these two tasks, this difference shows the impact of LLM inaccuracies on pretraining.

For \textbf{H2}, we test two different ways of using the pretrained models in the downstream rearrangement task. First, we directly finetune the pretrained model on the ground truth correct rearrangement; shown after the dashed vertical line in Figure \ref{fig:housekeep-ptft}. Here, the success rates for finetuned ELLM matches or outperform the baselines, especially if pretraining has already led to high success rates. Interestingly, we also find that the goal-conditioned ELLM variant consistently suffers a drop in performance when finetuning starts. We hypothesize this is due to the treatment of all suggested goals as a single string, so if any single goal changes between pretraining and finetuning the agent must relearn the goal embedding changes. Second, in Figure \ref{fig:housekeep-ft} we present results for directly training a new agent on the downstream task, using the frozen pretrained model as an exploratory actor during $\epsilon$-greedy exploration. Once again, we observe that ELLM consistently matches or outperforms all baselines. We also see here that the KB-IM baselines are more competitive, suggesting that this training scheme is better suited for pretrained exploration agents that are not well-aligned to the downstream task. 
\section{Conclusions and Discussion}

We have presented ELLM, an intrinsic motivation method that aims to bias exploration towards common-sense and plausibly useful behaviors via a pretrained LLM. We have shown that such priors are useful for pretraining agents in extrinsic-reward-free settings that require common-sense behaviors that other exploration methods fail to capture. 

ELLM goes beyond standard novelty search approaches by concentrating exploration on common-sensical goals. This is helpful in environments offering a wide array of possible behaviors among which very few can said to be \textit{plausibly useful}. It is less helpful in environments with little room for goal-based exploration, when human common-sense is irrelevant or cannot be expressed in language (\eg fine-grained manipulation), or where state information is not naturally encoded as a natural language string.

LLM performance is sensitive to prompt choice. Even with a well-chosen prompt, LLMs sometimes make errors, often due to missing domain-specific knowledge. False negatives can permanently prevent the agent from learning a key skill: in Crafter, for example, the LLM never suggests creating wood pickaxes. There are multiple avenues to address this limitation: (1)~combining ELLM rewards with other KB-IM rewards like RND, (2)~prompting LLMs with descriptions of past achievements (or other feedback about environment dynamics) so that LLMs can learn about the space of achievable goals, (3) injecting domain knowledge into LLM prompts, or (4) fine-tuning LLMs on task-specific data. While ELLM does not rely on this domain knowledge, when this information exists it is easy to incorporate. 

ELLM requires states and transition captions. Our learned captioner experiments Figure~\ref{fig:captioner} suggest we can learn these from human-labeled samples, but in some environments training this captioner might be less efficient than collecting demonstrations or hard-coding a reward function. Still, we are optimistic that as progress in general-purpose captioning models continues, off-the-shelf captioners will become feasible for more tasks. Lastly, suggestion quality improves considerably with model size. Querying massive LLMs regularly may be time- and cost-prohibitive in some RL environments.

As general-purpose generative models become available in domains other than text, ELLM-like approaches might also be used to suggest plausible visual goals, or goals in other state representations.
ELLM may thus serve as a platform for future work that develops even more general and flexible strategies for incorporating human background knowledge into reinforcement learning.

\section{Acknowledgements}
YD and OW are funded by the Center for Human-Compatible Artificial Intelligence.
CC received funding from the European Union’s Horizon 2020 research and innovation programme under the Marie Skłodowska-Curie grant agreement No. 101065949.
This material is based upon work supported by the National Science Foundation under Grant No.~2212310 to AG and JA. OpenAI credits for GPT-3 access were provided through OpenAI's Researcher Access Program. We thank Sam Toyer and the members of the RLL for feedback on early iterations of this project.

\bibliography{biblio}
\bibliographystyle{icml2023}

\newpage
\appendix
\onecolumn
\section{Crafter Pretraining Ablation} \label{sec:crafter_pretrain_all}

\begin{figure*}[h!]
\includegraphics[width=\columnwidth]{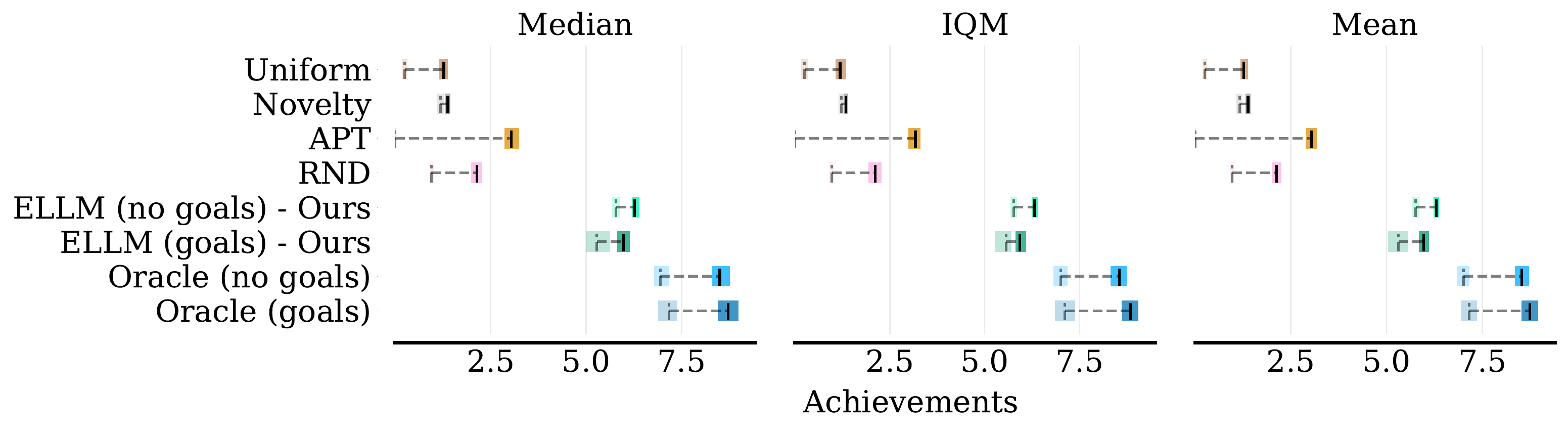}
\caption{Number of ground truth achievements unlocked per episode at the end of pretraining. We show the median, interquartile mean (IQM) and mean of the achievements measured in 10 evaluation trials, each averaged over 10 episodes and 5 seeds (50 points) \cite{agarwal2021deep}. Opaque bars represent variants leveraging textual observations in addition of visual ones and dashed lines represent the gap with vision-only variants (less opaque). We report results for each method described in Table~\ref{tab:baselines}. Results show that providing textual observations increases performance across all conditions.}
\label{fig:cpretrainingb}
\end{figure*}

\section{Crafter Downstream Training} \label{sec:crafter_finetune_full}

We finetune on seven downstream Crafter tasks plus the Crafter game reward:

\begin{itemize}
    \item \textbf{Place Crafting Table} - agent must chop a tree and then create a crafting table. This is an easy task most agents will have seen during pretraining.
    \item \textbf{Attack Cow} - agent must chase and attack a cow. This is also an easy task often seen during pretraining in most methods.
    \item \textbf{Make Wood Sword} - agent must chop a tree, use it to make a crafting table, chop a second tree, use the wood at the crafting table to make a wood sword. This task could be achieved during the pretraining env, but many agents rarely or never achieved it because of the sheer number of prerequisites.
    \item \textbf{Mine Stone} - agent must chop a tree, use it to make a crafting table, chop a second tree, use the wood at the crafting table to make a wood pickaxe, seek out stone, and then mine stone. This task is so challenging that we replaced the fully sparse reward (where all pretraining methods fail) with a semi-sparse reward for achieving each subtask.
    \item \textbf{Deforestation} - agent must chop 4 trees in a row. This task tests whether having goal conditioning improves performance by directing the agent. During pretraining most agents will have chopped a tree, but novelty bias should deter agents from regularly chopping 4 trees in a row.
    \item \textbf{Gardening} Like above, this task tests the value of goal conditioning. The agent must first collect water and then chop the grass. Both skills maybe have been learned during pretraining, but never in sequence.
    \item \textbf{Plant Row} - agent must plant two plants in a row. This task is challenging because even a highly skilled ELLM agent cannot have learned this task 0-shot because the state captioner has no concept of a ``row''.
\end{itemize}

\begin{figure*}[ht!]
\includegraphics[width=\textwidth]{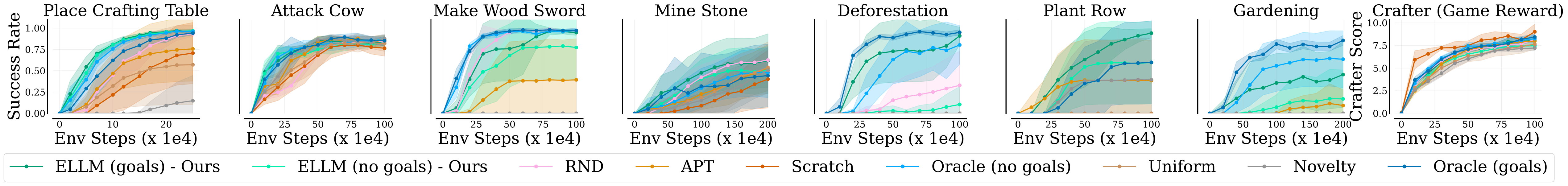}
\caption{Goal completion success rate for different tasks in the Crafter environment. RL training uses sparse rewards. Each method trains an agent from scratch while using a pretrained policy for exploration. Each line shows the mean across 5 seeds with shaded stds.}
\label{fig:crafter_finetune_full}
\end{figure*} 

\section{Crafter Env Modifications} \label{sec:crafter_action_space}

\begin{figure*}[ht!]
\includegraphics[width=\columnwidth]{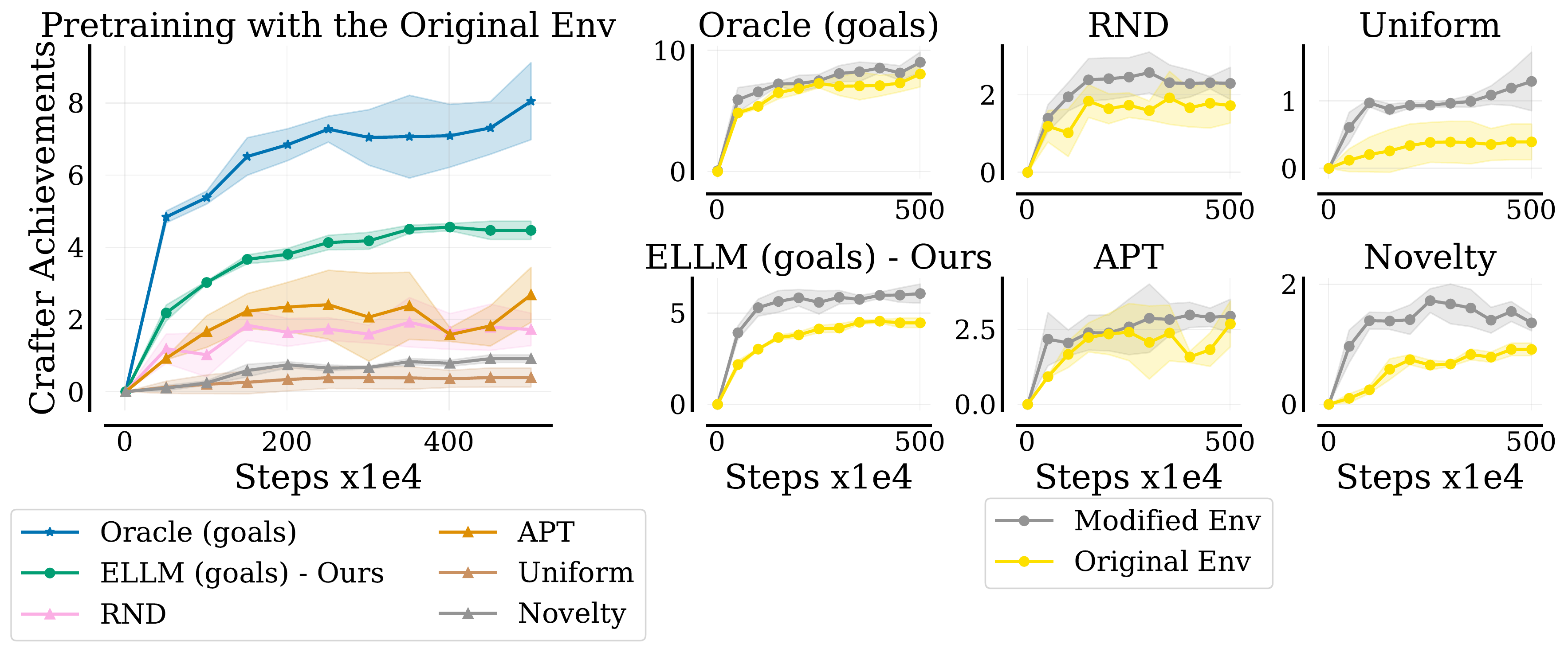}
\caption{Training without the environment simplifications described in Section \ref{sec:crafter}.  Left: pretraining results (comparable to Figure \ref{fig:cpretraininga}). Right: original vs modified env performance. Curves average over 3 seeds with std shading. We see minor performance changes across most algorithms but no change in the rank-order of methods.}
\label{fig:crafter_original}
\end{figure*}

The default Crafter action space contains an all purpose ``do'' action which takes different actions depending on what object the agent is facing - for instance attacking a skeleton, chopping a tree, or drinking water.

We modify the action space to increase the exploration problem by turning the general `do' action into more precise combinations of action verbs + noun arguments. Whereas `do' previously was an all purpose action that could attack a skeleton, chop a tree, or drink water, the agent must now learn to choose between the actions as arbitrary verb + noun combinations, `\texttt{attack skeleton}', `\texttt{chop tree}', `\texttt{drink water}.' The exploration problem becomes more difficult as this larger combinatorial action space is not restricted to admissible actions and the agent could try to \texttt{drink skeleton} or \texttt{attack water}. Whereas the old action space was 17-dimensional, our new combinatorial one contains 260 possible actions. One way to impose human priors is to design the agent's action space explicitly to disallow invalid combinations (\eg \texttt{'drink' + 'furnace'}). However, manually designing and imposing such constraints is also unlikely to be scalable. We hypothesize that our method, guided by common-sense knowledge from LLMs, will focus on learning to use only meaningful action combinations. For the purposes of the Novelty and Uniform baselines, which reward agents for achieving even nonsensical goals, we consider a goal ``achieved'' if the agent takes an action in front of the appropriate target object (e.g taking ``drink furnace'' in front of a furnace).

\section{Crafter Prompt} \label{sec:crafter-prompt}

\texttt{Valid actions: sleep, eat, attack, chop, drink, place, make, mine}

\texttt{You are a player playing a game. Suggest the best actions the player can take based on the things you see and the items in your inventory. Only use valid actions and objects.}

\texttt{You see plant, tree, and skeleton. You are targeting skeleton. What do you do?}

\texttt{- Eat plant}

\texttt{- Chop tree}

\texttt{- Attack skeleton}

\texttt{You see water, grass, cow, and diamond. You are targeting grass. You have in your inventory plant. What do you do?}

\texttt{- Drink water}

\texttt{- Chop grass}

\texttt{- Attack cow}

\texttt{- Place plant}

In total, the actions present in the prompt make up:
\begin{itemize}
    \item 6 / 10 (60\%) of the good actions the ELLM agent receives.
    \item 6 / 21 (28.6\%) of all rewarded actions the agent receives.
    \item 7 / 15 (50\%) of all good action suggested.
    \item 7 / 51 (13.7\%) of all actions suggested.
\end{itemize}

In future work, it would be interesting to explore how performance changes with fewer actions included in the prompt. \rebuttal{As a preliminary experiment, we have found that pretraining performance is maintained if you provide a prompt with only one example of a list of valid goals. The list only contains two goals. Instead, we use more extensive instructions to tell the agent what good suggestions look like.  See the prompt below and pretraining comparison in Figure \ref{fig:new_prompt}. This new prompt comes with a decrease in the fraction of ``Good'' suggestions (shown in Table \ref{tab:new_prompt}, showing that suggestion accuracy is not perfectly correlated with success.

New prompt:
\texttt{Valid actions: sleep, eat, attack, chop, drink, place, make, mine}

\texttt{You are a player playing a Minecraft-like game. Suggest the best actions the player can take according to the following instructions.}

\texttt{1. Make suggestions based on the things you see and the items in your inventory.}

\texttt{2. Each scene is independent. Only make suggestions based on the visible objects, status, and inventory in the current scene.}

\texttt{3. Each suggestion should either be a single valid action, or a phrase consisting of an action and an object. (example: "Eat plant").}

\texttt{4. Do not make suggestions which are not possible or not desirable, such as ``Eat skeleton''.}

\texttt{5. Only make suggestions which are reasonable given the current scene (e.g. only ``Eat plant'' if a plant is visible).}

\texttt{6. You may suggest multiple actions with the same object, but do not duplicate list items.}

\texttt{7. Use your knowledge of Minecraft to make suggestions.}

\texttt{8. Prioritize actions which involve the object you are facing or which the agent hasn't achieved before.}

\texttt{9. Each scene will include a minimum and maximum number of suggestions. Stick within this range.}

\texttt{New scene: You see plant, cow, and skeleton. You are facing skeleton. What do you do (include 1-2 suggestions)?}

\texttt{- Eat plant}

\texttt{- Attack skeleton}
\\

\texttt{New scene: You see [INSERT CURRENT SCENE DESCRIPTION.] What do you do (include 2-7 suggestions)?}

\begin{table}[]
    \centering
    \footnotesize
    \begin{tabular}{lcc}
    \toprule
    & Suggested & Rewarded  \\
    \midrule
       Context-Insensitive & 21.0\% &  0.8\%  \\ 
       Common-Sense Insensitive & 20.5\%  & 54.8\%   \\   
       Good & 34.1\%  & 44.4\%   \\ 
       Impossible & 24.5\% & 0\%  \\
       \bottomrule
    \end{tabular}
    \caption{\rebuttal{Fractions of suggested and rewarded goals which are good, generated with the modified two-example prompt.}}
    \label{tab:new_prompt}
\end{table}

\begin{figure}[]
\centering
\includegraphics[width=.75\columnwidth]{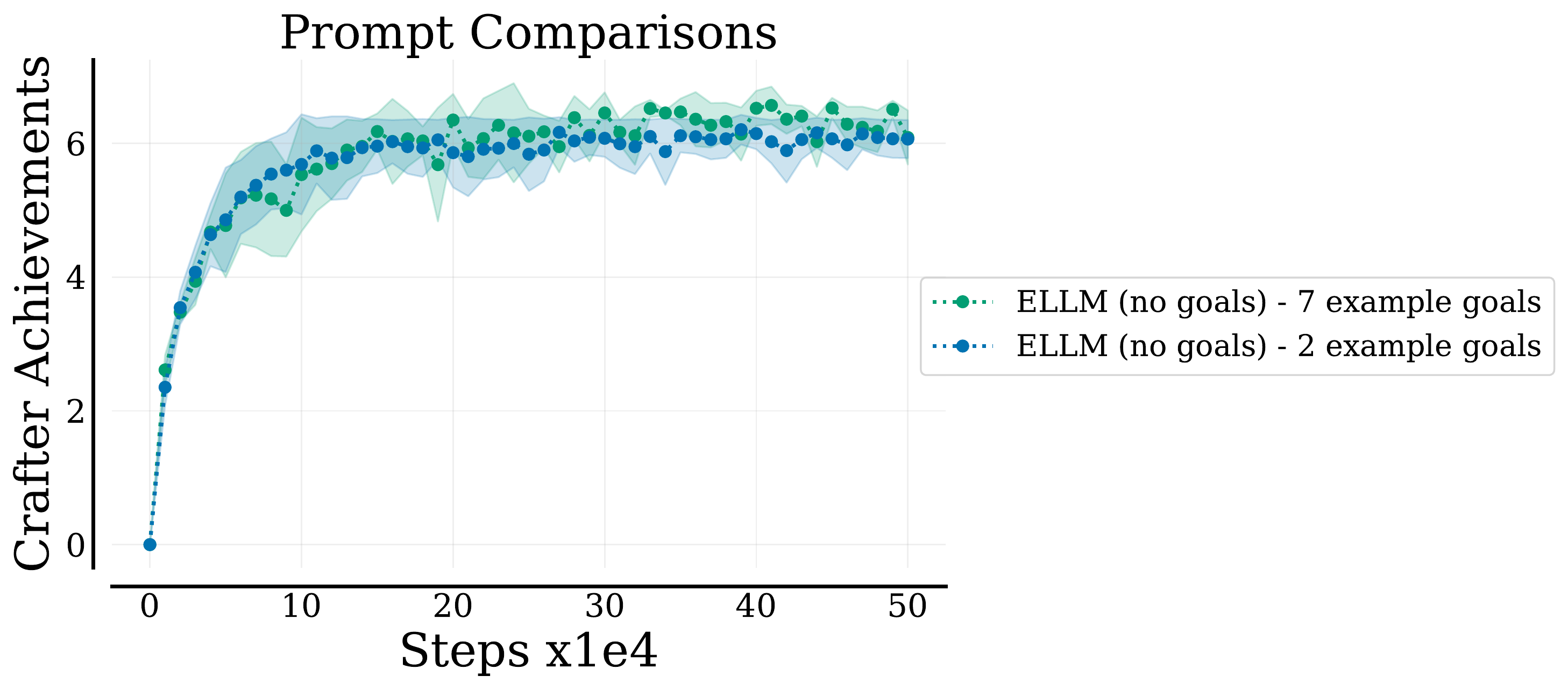}
\caption{\rebuttal{Comparison between performance of the prompt containing 7 suggested goals (used throughout the paper) and a modified prompt which only includes 2 examples.}}
\label{fig:new_prompt}
\end{figure}
}

\section{Crafter Action Space}
\rebuttal{
We expand the action space of Crafter to increase exploration difficulty and study if ELLM can learn to avoid nonsensical or infeasible actions. The full action space consists of just verbs (for actions that do not act on anything, such as \texttt{sleep}) or verb + noun combinations as follows:

\begin{itemize}
    \item Verbs: \texttt{do nothing} (no noun), \texttt{move left} (no noun), \texttt{move right} (no noun), \texttt{move up} (no noun), \texttt{move down} (no noun), \texttt{sleep} (no noun), \texttt{mine}, \texttt{eat}, \texttt{attack}, \texttt{chop}, \texttt{drink}, \texttt{place}, \texttt{make}
    \item Nouns: \texttt{zombie, skeleton, cow, tree, stone, coal, iron, diamond, water, grass, crafting table, furnace, plant, wood pickaxe, stone pickaxe, iron pickaxe, wood sword, stone sword, iron sword}
\end{itemize}

For example, an action can be \texttt{drink water} or \texttt{drink grass}.
}

\section{Housekeep Tasks} \label{sec:housekeep-tasks}
The original Housekeep benchmark features a large set of different household scenes and episodes with different objects and receptacles possibly instantiated. The ground truth correct object-receptacle placements were determined by crowdsourcing humans.  However, since our focus is on RL pretraining, we do not make use of the mapping and planning methods from the original benchmark. To scope the problem for RL, we focus on the first 4 tasks with 5 different misplaced objects per task. 

\begin{table}[ht!]
    \centering
    \begin{tabular}{c|c}
         & Misplaced Objects  \\ \hline
        Task 1 & peppermint, lamp, lantern, herring fillets, vase \\ 
        Task 2 & lamp, sparkling water, plant, candle holder, mustard bottle\\ 
        Task 3 & pepsi can pack, electric heater, helmet, golf ball, fruit snack\\ 
        Task 4 & chocolate, ramekin, pan, shredder, knife\\ 
    \end{tabular}
    \caption{Objects per task}
    \label{tab:my_label}
\end{table}

\section{Housekeep Prompt} \label{sec:housekeep-prompt}

\texttt{You are a robot in a house. You have the ability to pick up objects and place them in new locations. For each example, state if the item should be stored in/on the receptacle.}

\texttt{Should you store a dirty spoon in/on the chair: No.}

\texttt{Should you store a mixing bowl in/on the dishwasher: Yes.}

\texttt{Should you store a clean sock in/on the drawer: Yes.}

\section{Algorithmic Details}\label{sec:alg-details}

We make use of DQN \cite{mnih2013playing}, with double Q-learning \cite{van2016deep}, dueling networks \cite{wang2016dueling}, and multi-step learning \cite{sutton1998introduction}. 

\begin{table}[]
    \centering
    \begin{tabular}{c|c|c}
       Name  & Value (Crafter) & Value (Housekeep)  \\ \hline
        Frame Stack & 4  &4 \\
        $\gamma$ & .99 & .99 \\
        Seed Frames & 5000 & 5000 \\\
        $n$-step & 3 & 3 \\
        batch size & 64 & 256 \\
        lr & 6.25e-5 & 1e-4 \\ 
        target update $\tau$ & 1.0 & 1.0 \\
        $\epsilon$-min & 0.01 & 0.1 \\ 
        update frequency & 4 & 4 
    \end{tabular}
    \caption{DQN Hyperparameters}
    \label{tab:my_label}
\end{table}

For both environments, policies take in 84 $\times$ 84 images which are encoded using the standard Nature Atari CNN \cite{mnih2015human}. The image is then passed through a linear layer to output a 512 dimensional vector. If the policy is text-conditioned, we compute the language embedding of the state caption using \texttt{paraphrase-MiniLM-L3-v2} SBERT model \cite{reimers-2019-sentence-bert}, and if the policy is goal-conditioned we similarly compute the language embedding of the goals $g_{1:k}$ using \texttt{paraphrase-MiniLM-L3-v2}. \rebuttal{We encode all goals as a single text sequence as we did not see any improvement from encoding them each separately and summing or concatenating the embeddings.} The image and text embeddings are then concatenated together before being passed to the Q-networks. Each of the value and advantage streams of the Q-function are parametrized as 3-layer MLPs, with hidden dimensions of 512 and ReLU nonlinearities.  

In the Crafter environment, we swept over the following hyperparameters for the Oracle and Scratch (no-pretraining) conditions: learning rate, exploration decay schedule, and network update frequency. We then applied these hyperparameters to all conditions, after confirming that the hyperparameters were broadly successful in each case.

For Housekeep pretraining, we swept lr $\in [1e-3, 1e-4, 1e-5]$, $\epsilon$-min $\in [0.1, 0.01]$, and batch size $\in [64, 256]$.

\section{Hard-coded Captioner Details}
\rebuttal{
\paragraph{Crafter} The state captioner is based on the template shown in Figure \ref{fig:example_caption} (left). This consists of three components: the observation, the items, and the agent status.
\begin{itemize}
    \item Observation: We take the underlying semantic representation of the current image from the simulator. Essentially this maps each visible grid cell to a text description (e.g. each tree graphic is mapped to “tree”). We then take this set of descriptions (i.e. not accounting for the number of each object) and populate the “observation” cell of the template.
    \item Items: We convert each of the inventory items to the corresponding text descriptor, and use this set of descriptions to populate the “item” cell of the template. 
    \item Health status: We check if any of the health statuses are below maximum, and if so, convert each to a corresponding language description (e.g. if the hunger status is $<9$, we say the agent is “hungry”).
\end{itemize}
The transition captioner uses the action labels. Each action maps to a predefined verb + noun pairing directly (e.g. “eat cow”).

\paragraph{Housekeep}
The state captioner is based on the template shown in Figure \ref{fig:example_caption} (right). We use the simulator’s semantic sensor to get a list of all visible objects, receptacles, and the currently held object. The transition captioner is also based on the simulator’s semantic sensor, which indicates which receptacles the visible objects are currently in. }

\section{Learned Crafter Captioner} \label{sec:captioner_appendix}

The captioner is trained with a slightly modified ClipCap algorithm \citep{clipcap} on a dataset of trajectories generated by a trained policy using the PPO implementation from \citet{stanic2022learning}. Visual observations at timestep $t$ and $t+1$ are embedded with a pretrained and frozen CLIP ViT-B-32 model \citep{radford2021learning} and concatenated together with the difference in semantic embeddings between the two corresponding states. Semantic embeddings include the inventory and a multi-hot embedding of the set of objects present in the local view of the agent. This concatenated representation of the transition is then mapped through a learned mapping function to a sequence of 10 tokens. Finally, we use these 10 tokens as a prefix and pursue decoding using a pretrained and frozen GPT-2 to generate the caption \citep{radford2019language}. We train the mapping from transition representation to GPT tokens on a dataset of 847 human labels and 900 synthetic labels obtained by sampling from a set of between 3 and 8 different captions for each each distinct type of transitions. Instead of the programmatic ``\texttt{chop tree}'' and ``\texttt{attack zombie},'' labeled captions involve fully-formed sentences: ``\texttt{You collected a sapling from the ground},'' ``\texttt{You built a sword out of wood},'' or ``\texttt{You just stared at the sea}.'' Because of this additional linguistic diversity, we compare captions to goals with a lower cosine similarity threshold of .5.

\begin{figure*}[ht!]
\centering
\includegraphics[width=0.5\textwidth]{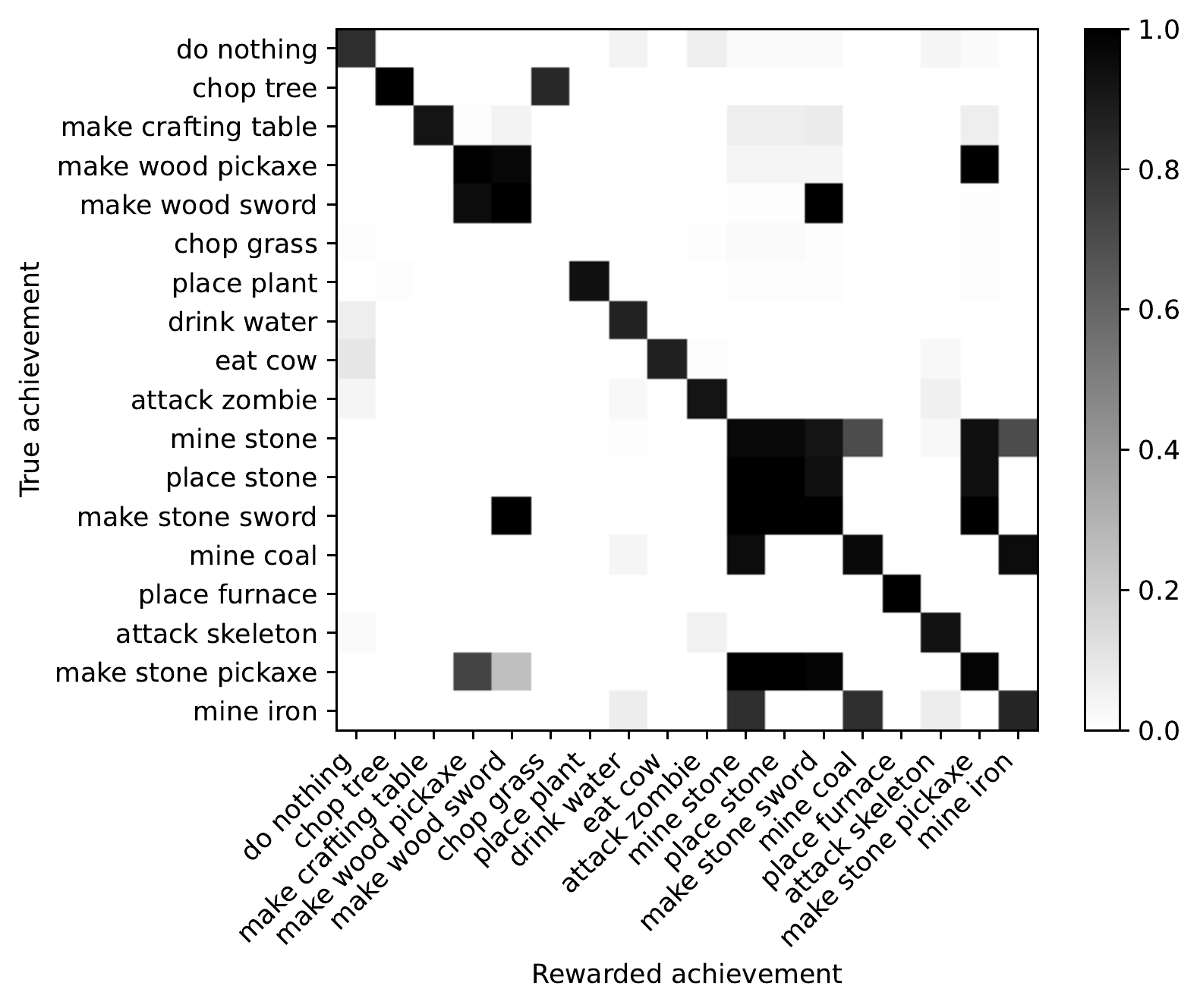}
\caption{Reward confusion matrix. Each row gives the probability that any of the column achievement is detected when the row achievement is truly unlocked. For instance, in row 2, when the agent chops a tree, with high probability the agent will be rewarded for the ``chop tree'' and ``chop grass'' actions. Tested on trajectories collected from an expert PPO policy, each row estimates probabilities using between 27 and 100 datapoints (27 for \textit{mine iron}, the rarest achievements). Rows do not sum to one, as a given achievement, depending on its particular caption, could potentially trigger several rewards.}
\label{fig:conf_mat_captioner}
\end{figure*}

Imperfect captioners can cause learning issues in two different ways: (1)~they can generate wrong captions all together and (2)~they can generate a valid caption that still lead to faulty reward computations. If the caption is linguistically too different from the achievement it captions, the similarity-based reward might not be able to pick it up (false negative reward). This same linguistic variability might cause the reward function to detect the achievement of another achievement that was not achieved (false positive reward). Figure~\ref{fig:conf_mat_captioner} measures all these issues at once. For each row, it answers: what is the probability that the reward function would detect a positive reward for each of the column achievements when the true achievement is the row label? The false negative rate is 11\% on average (1 - the diagonal values), with a much higher false negative rate for \textit{chop grass} (100\%). Indeed, human caption mentioned the outcome of that action instead of the action itself (\textit{collect sapling}); which the similarity-based reward fails to capture. The false positive rate (all non diagonal values) is significant here: the agent can get rewarded for several achievements it did not unlock. This often occurs when achievements share words (\eg wood, stone, collect). This indicates a difficulty of the semantic similarity to differentiate between achievements involving these words.

\section{Crafter LLM Analysis} \label{sec:crafter_llm}

Table \ref{tab:crafter_lm_breakdown} shows that the actions agents are rewarded for are dominated by good actions (66.5\%) and bad actions (32.4\%). This makes sense; impossible actions can never be achieved. Most context-insensitive cannot be achieved (e.g. ``drink water'' suggested when no water is present). We consider an action a ``success'' by checking whether the agent attempted a particular action in front of the right object, so the agent occasionally is rewarded when it takes a context-insensitive action in the appropriate physical location but without the necessary prerequisites (e.g. mining stone without a pickaxe).

Table \ref{tab:crafter_llm_examples} gives examples of LLM suggestions in Crafter. 

\begin{table}[ht!]
    \centering
    
    \footnotesize
    \begin{tabular}{lcccc}
    \toprule
    \textbf{Suggestion Type} & \textbf{Examples} \\
    \midrule
       \textbf{Good} & chop tree, attack skeleton, place plant  \\ 
       \textbf{Context-Insensitive} & make crafting table (without wood), mine stone (without a pickaxe or not by stone) \\ 
       \textbf{Common-Sense-Insensitive} & mine grass, make diamond, attack plant \\ 
       \textbf{Impossible} & make path, make wood, place lava \\ 
       \bottomrule
    \end{tabular}
    \caption{Classification accuracy of LLM for each Housekeep task (left column is true positives, right column is true negatives).}
    \label{tab:crafter_llm_examples}
\end{table}

\rebuttal{
\section{Novelty Bonus Ablation} \label{sec:novelty-ablation}

We ablate the importance of ELLM's novelty bias in Figure \ref{fig:crafter-novelty-ablation} by allowing the agent to be rewarded repeatedly for achieving the same goal. We see that without the novelty bonus the agent only learns to repeat a small set of easy goals and fails to explore diversely.

\begin{figure}
\centering
\begin{subfigure}{.5\linewidth}
\includegraphics[width=.9\columnwidth]{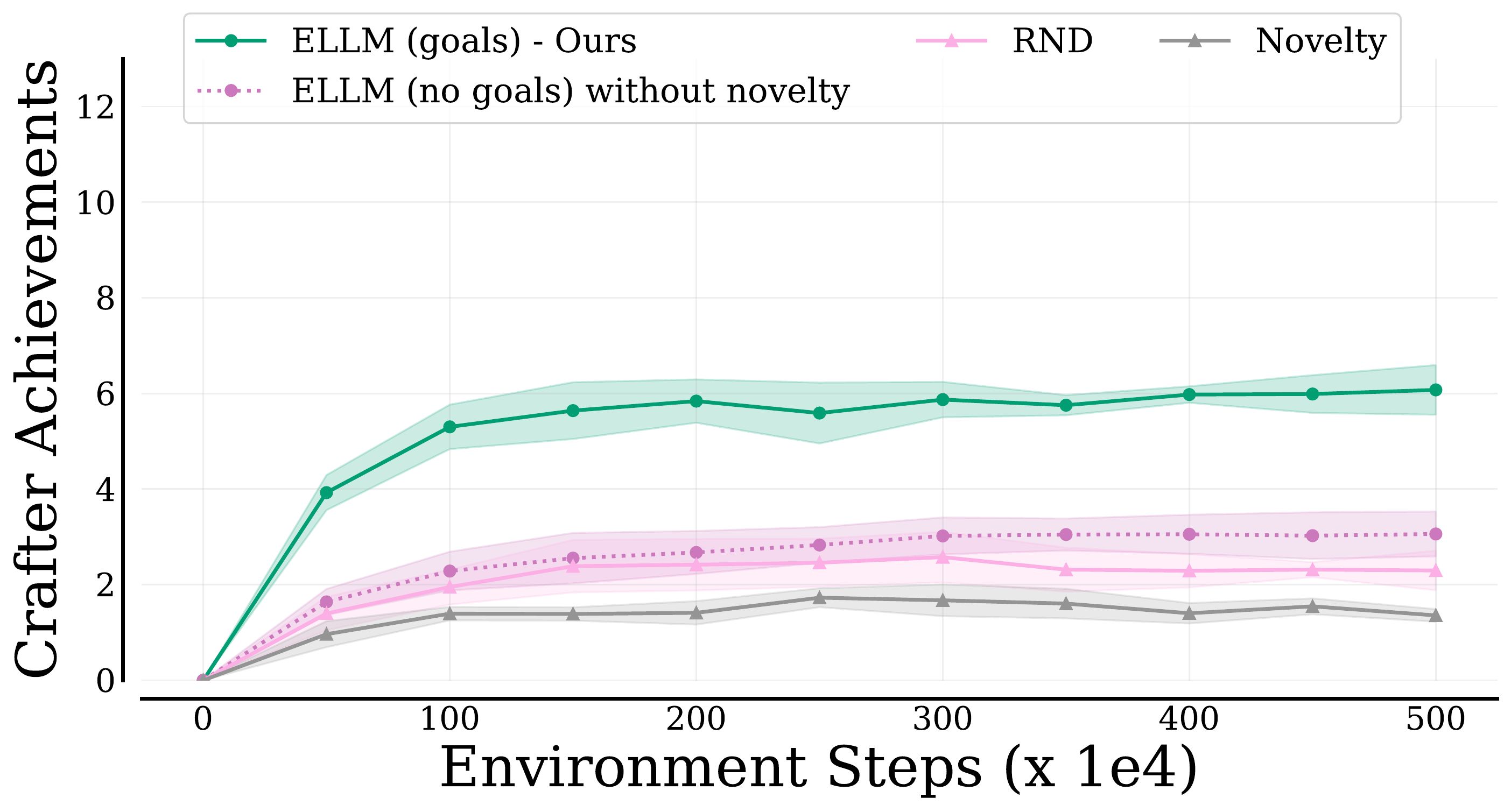}
\caption{\rebuttal{Crafter pretraining runs (similar to Figure \ref{fig:cpretraininga}), including the ``ELLM without novelty'' ablation where ELLM's novelty bias is removed.}}
\end{subfigure}\\
\begin{subfigure}{.8\linewidth}
\includegraphics[width=\textwidth]{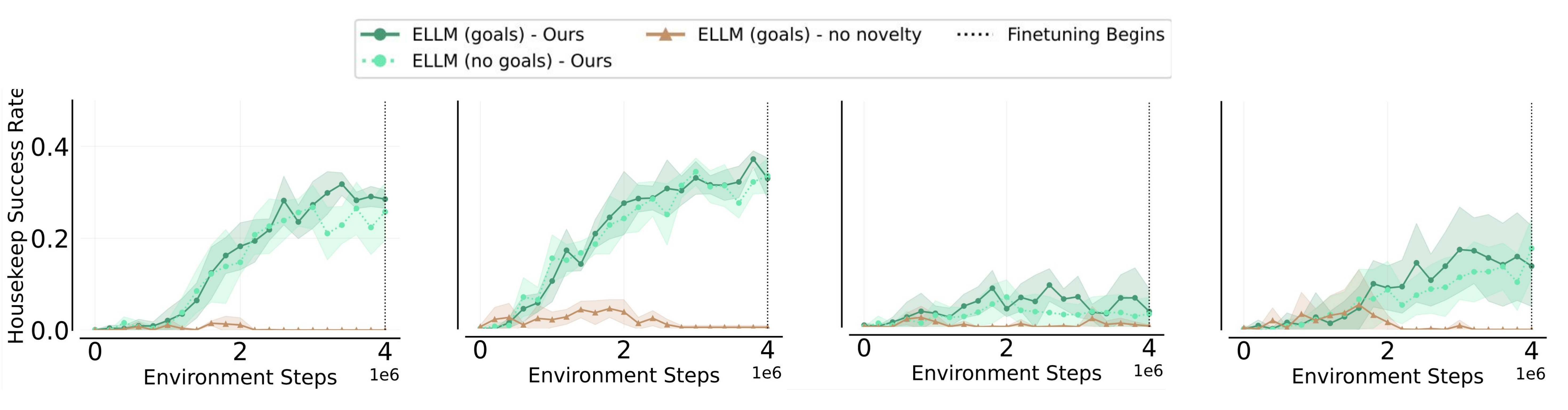}
\caption{\rebuttal{Housekeep pretraining runs (similar to Figure \ref{fig:housekeep-ptft}), including the ``ELLM without novelty'' ablation where ELLM's novelty bias is removed.}}
\end{subfigure}
\caption{}
\label{fig:crafter-novelty-ablation}
\end{figure}
}

\rebuttal{
\section{Analysis of Downstream Training Approaches} \label{sec:expl_vs_finetune}

We explored two methods for using exploratory policies: \textit{finetuning}, where the weights of the exploration policy are finetuned and the \textit{guided exploration} method, where a new policy is trained from scratch and the pretrained policy is used for $\epsilon$-greedy exploration. 

We found that in Housekeep both methods are effective for ELLM (Figure \ref{fig:housekeep-ptft} and Figure \ref{fig:housekeep-ft}). However, in Crafter we found that the finetuning method performed poorly across all methods (ELLM, baselines, and oracles). Often, we observed that early in finetuning, the agent would unlearn all of its previous useful behaviors, including moving around the environment to interact with objects. We hypothesize that this due to a mismatch in the density and magnitude of rewards between pretraining and finetuning. When the finetuning agent finds it is achieving much lower than the expected return for taking its typical actions, it down-weights the likelihood of taking those actions and unlearns its previous skills. We found that decreasing the learning rate, freezing early layers of the network, manually adjusting finetuning rewards to be at the same scale as pretraining rewards, and decreasing the initial exploration rate partially mitigated this problem. However, these also decrease the sample efficiency and/or performance at convergence of the finetuned policy compared to a training-from-scratch baseline. In Figure \ref{fig:crafter-finetune-compare}), across all methods this method is less reliable than the guided exploration method (Figure \ref{fig:crafter_finetune}).

These findings are consistent with our Housekeep findings. In that environment, the ELLM pretraining task (achieving object placements suggested by a LLM) and the finetuning task (achieving object placements suggested by humans) are similar enough we only see minor dips in performance when finetuning starts. However, the RND and APT baselines have a greater pretrain-finetune mismatch, and we observe those methods did comparatively better with the guided exploration method.

\begin{figure*}[ht!]
\includegraphics[width=\textwidth]{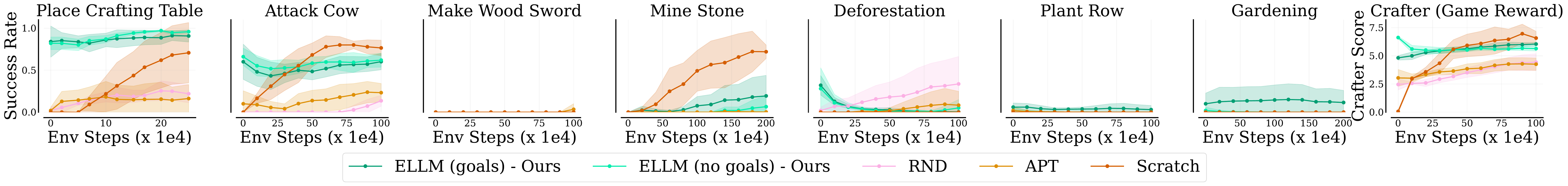}
\caption{\rebuttal{Success rates across training for each of the seven downstream tasks in the Crafter environment. Each run finetunes the pretrained agent using a lower learning rate than used during pretraining ($2e-5$). Plots show mean $\pm$ std for 5 seeds}}
\label{fig:crafter-finetune-compare}
\end{figure*} 
}

\section{Additional Baselines} \label{sec:noveld}

\rebuttal{We also include experiments with NovelD \cite{zhang2021noveld} in Figure \ref{fig:noveld}, a state-of-the-art exploration method which uses an estimate of state novelty to reward the agent for moving to more novel states. During pretraining, we find it performs similarly to the other prior-free intrinsic motivation methods.

\begin{figure}
\centering
\begin{subfigure}{.5\linewidth}
    \includegraphics[width=.9\columnwidth]{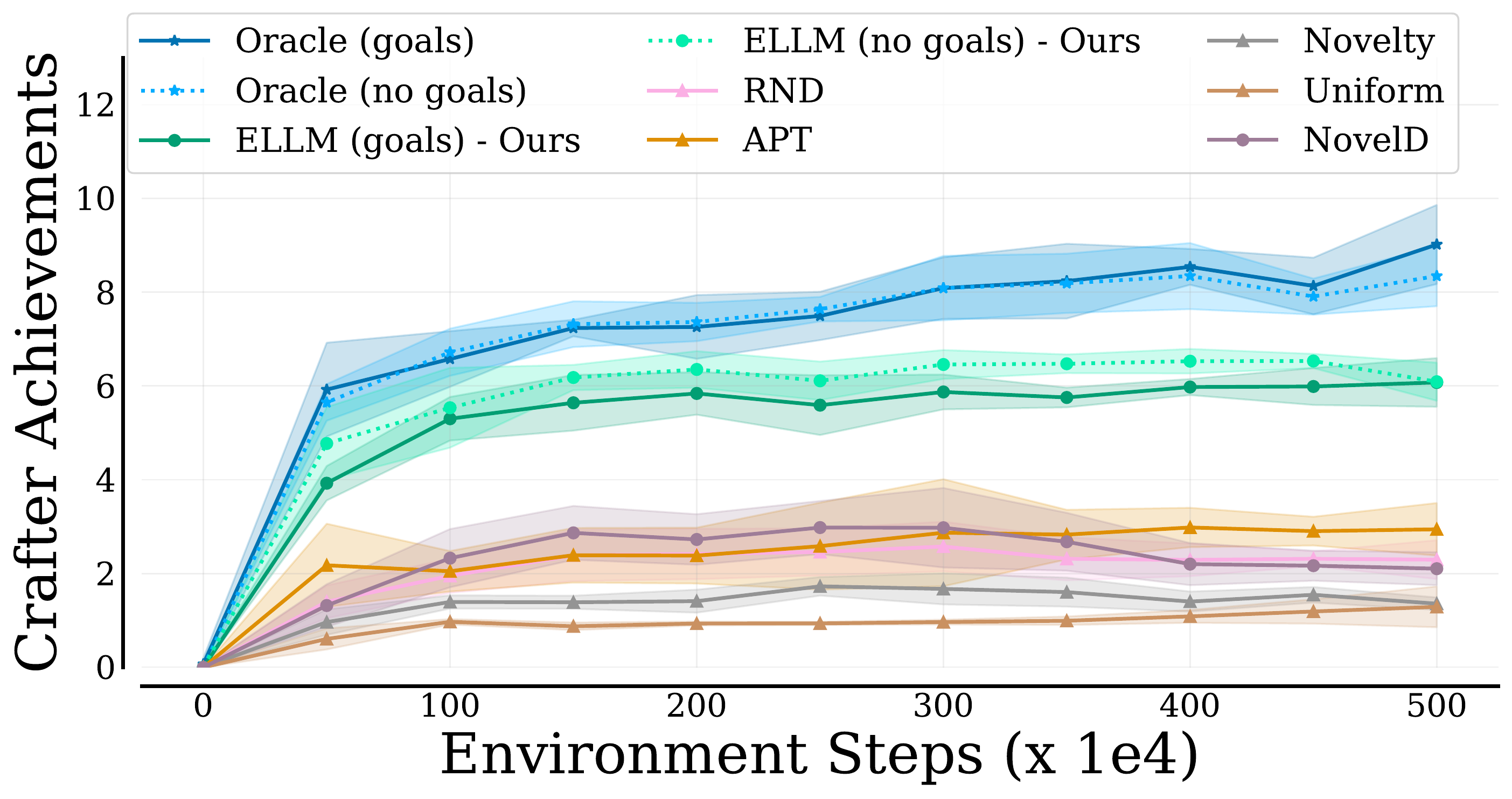}
    \caption{Crafter pretraining curve as in Figure \ref{fig:cpretraininga}, including NovelD baseline}
\end{subfigure}\\
\begin{subfigure}{\linewidth}
    \includegraphics[width=\columnwidth]{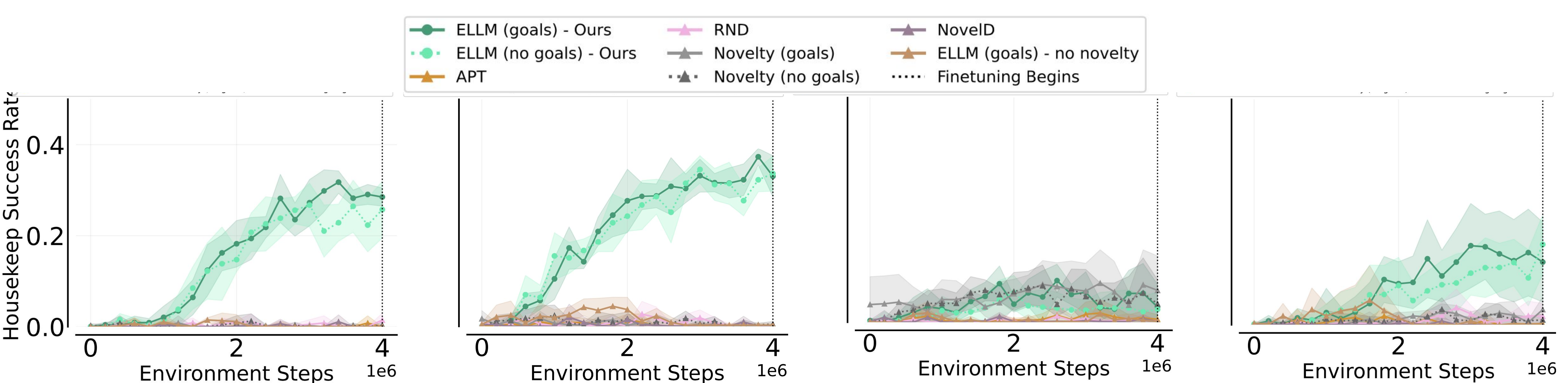}
    \caption{Housekeep pretraining curves as in Figure \ref{fig:housekeep-ptft}, including NovelD baseline}
\end{subfigure}
    \caption{Additional pretraining curves including NovelD.}
    \label{fig:noveld}
\end{figure}

}

\section{Code and Compute}

All code will be released soon, licensed under the MIT license (with Crafter, Housekeep licensed under their respective licenses). 

For LLM access, we use OpenAI's APIs. Initial experiments with the smaller GPT-3 models led to degraded performance, hence choosing Codex and Davinci for our experiments. Codex is free to use  and Davinci is priced at \$0.02/1000 tokens. We find caching to be significantly helpful in reducing the number of queries made to the API. Each API query takes .02 seconds, so without caching a single 5-million step training run would spend 27 hours querying the API (and far more once we hit the OpenAI rate limit) and cost thousands of dollars. Since we cache heavily and reuse the cache across runs, by the end of our experimentation, were make almost no API queries per run.

We use NVIDIA TITAN Xps and NVIDIA GeForce RTX 2080 Tis, with 2-3 seeds per GPU and running at roughtly 100ksteps/hour. Across all the ablations, this amounts to approximately 100 GPUs for pretraining.

\section{Societal Impact} 
While LLMs priors have been shown to exhibit impressive common-sense capabilities, it is also well-known that such models are highly prone to harmful social biases and stereotypes \citep{bender2021dangers, abid2021persistent,nadeem2020stereoset}. When using such models as reward functions for RL, as in ELLM, it is necessary to fully understand and mitigate any possible negative behaviors that can be learned as a result of such biases. While we focus on simulated environments and tasks in this work, we emphasize that more careful study is necessary if such a system is deployed to more open-ended learning in the real world. Potential mitigations with ELLM specifically can be: actively filtering LLM generations for harmful content before using them as suggested goals, prompting the LM with guidelines about what kinds of prompts to output, and/or using only the closed-form ELLM variant with more carefully constrained goal spaces.



\end{document}